\RequirePackage[svgnames,table]{xcolor}

\documentclass[11pt,letterpaper,logo]{yalearxiv}

\usepackage{graphicx}
\usepackage{float,epstopdf}
\usepackage{bbm}

\usepackage{microtype}

\usepackage{natbib}
\setcitestyle{square}

\usepackage{subcaption}
\usepackage{booktabs}

\usepackage{amsmath}
\usepackage{amssymb}
\usepackage{mathtools}
\usepackage{amsthm}
\usepackage{dsfont}
\usepackage{multicol}
\usepackage{makecell}
\usepackage{multirow} 
\usepackage{amsfonts} 
\usepackage{mathrsfs}
\usepackage[amssymb, thickqspace]{SIunits}
\usepackage{enumitem}
\usepackage{pgfplotstable}
\usepackage{lipsum}		

\usepackage{microtype}
\usepackage{graphicx}
\usepackage{booktabs} 
\usepackage[table]{xcolor}
\usepackage{arydshln}
\usepackage[normalem]{ulem} 

\usepackage{cases}
\usepackage{wrapfig}

\usepackage{url}

\usepackage{thmtools}
\usepackage{thm-restate}
\usepackage{tabu}

\definecolor{huskypurple}{HTML}{4B2E83}

\usepackage{titletoc}

\usepackage{listings}
\lstdefinestyle{promptstyle}{
  basicstyle=\ttfamily\footnotesize,
  breaklines=true,
  breakautoindent=false,
  breakindent=0pt,
  postbreak=\mbox{\textcolor{gray}{$\hookrightarrow$}\space},
  columns=fullflexible,
  keepspaces=true,
  frame=single,
  framesep=5pt,
  xleftmargin=6pt,
  xrightmargin=6pt,
  aboveskip=8pt,
  belowskip=8pt,
  showstringspaces=false,
}   
\makeatletter
\def\munderbar#1{\underline{\sbox\tw@{$#1$}\dp\tw@\z@\box\tw@}}
\makeatother

\AddToHook{cmd/appendix/before}{%
  \setcounter{axiom}{0}%
}

\newcommand{\be}{\begin{equation}}
\newcommand{\ee}{\end{equation}}
\newcommand{\bea}{\begin{equation*}\begin{aligned}}
\newcommand{\eea}{\end{aligned}\end{equation*}}

\newtcolorbox{simpleElegantQuote}{
    colback=AliceBlue!50!White,
    colframe=RoyalBlue!75!Black,
    boxrule=0.5pt,
    arc=2mm,
    boxsep=4pt,
    left=10pt, right=10pt,
    top=8pt, bottom=8pt,
    fontupper=\itshape,
}

\title{Can Code Carry Artistic Style?\\ Executable Programs as Transferable Art}

\runningtitle{CodeArtist --- Technical Report}


\usepackage{fontawesome5}   

\definecolor{yaleblue}{RGB}{0,58,112}

\author{
    Siru Jiang\textsuperscript{1} \quad 
    Yongzhe Lyu\textsuperscript{2} \quad 
    Shuo Lu\textsuperscript{1} \quad 
    Yubin Wang\textsuperscript{3} \quad 
    Jian Liang\textsuperscript{1,\(\dagger\)} \\
    \textsuperscript{1}NLPR \& MAIS, CASIA \quad 
    \textsuperscript{2}PKU \quad 
    \textsuperscript{3}HKUST \\
    \textsuperscript{\(\dagger\)}Corresponding author. \\
    \faGithub~\textbf{Source Code:} \href{https://github.com/sirujiang/CodeArtist}{\texttt{https://github.com/sirujiang/CodeArtist}}
}
\hypersetup{colorlinks=true, linkcolor=blue!50!black, citecolor=blue!50!black,
            urlcolor=blue!50!black}

\usepackage{graphicx}
\usepackage{subcaption}
\usepackage{caption}

\begin{document}


\begin{abstract}
Coding agents are increasingly adopted for visual content generation, with growing interest in enhancing them with expressive artistic styles. 
However, existing style transfer skills mainly rely on textual instructions, requiring agents to re-implement low-level drawing details for each new scene and leading to inconsistent outputs.
Meanwhile, diffusion-based methods offer high visual quality and reusable style adaptation, yet their implicit representations make it difficult to maintain cross-scene consistency and make fine-grained editing.
Inspired by procedural painterly rendering and reusable program libraries, we explore whether artistic styles can be distilled into \textbf{reusable}, \textbf{transferable}, and \textbf{improvable} programs. 
In this work, we introduce \textbf{\textit{CodeArtist}}, a Python-based painting engine that encapsulates artistic styles, including palettes, brushwork, and textures, into reusable modules while decoupling them from scene content and motion. 
Once implemented as executable code, a style can be directly adapted to new scenes without revisiting original artwork, shared from frontier to lower-cost models, and incrementally refined as foundation models advance.
Experiments across diverse scenes and models demonstrate consistent style preservation, effective cross-model sharing, and cumulative improvements through modular updates.
Overall, we take a preliminary step toward exploring code as transferable art, offering new insights into how aesthetics can persist, be shared, and evolve across models and tasks.
\end{abstract}
\maketitle

\begin{center}
    \includegraphics[width=0.8\linewidth]{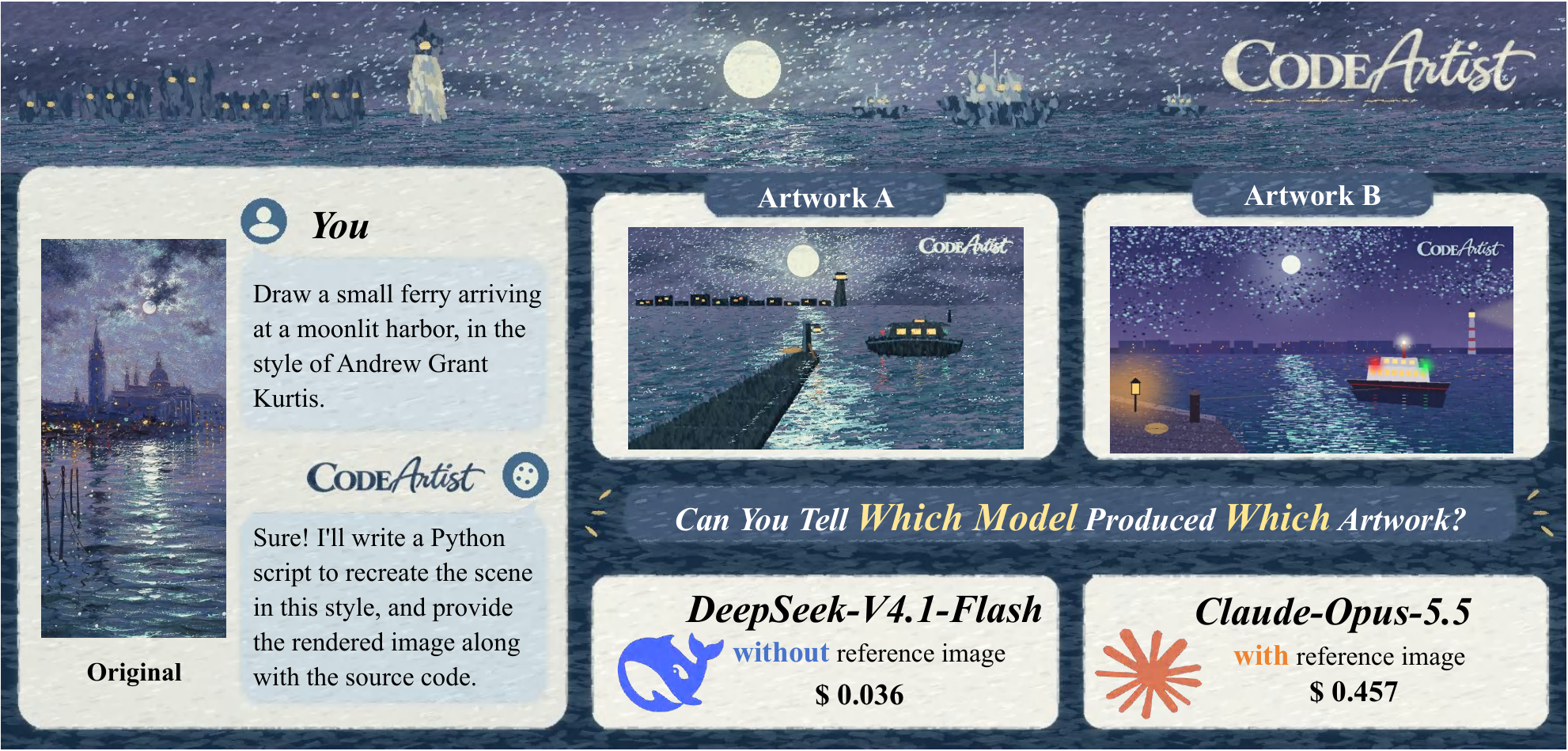}
    \captionof{figure}{\textbf{A blind test.}
    One artwork is generated without access to the reference image, while the other is created directly from the reference. 
    The teaser itself is also produced with \textit{CodeArtist}, and its step-by-step refinement is presented in our experiments.}
\end{center}

\section{Introduction}
As coding agents become increasingly capable, they are being used to generate a growing range of visual content through executable programs. 
Recent code-for-video systems can organize educational content, construct animated scenes, and produce diverse videos~\citep{code2video2025chen,vode2videocompass2026chen,vibevideobench2026diao}, highlighting the advantages of code in organizing content, arranging spatial layouts, controlling temporal dynamics, and structuring narratives.
However, these works primarily focus on structural fidelity and semantic alignment, overlooking artistic style despite its central role in how videos are perceived.
Existing approaches struggle to make artistic style persistent and controllable along two fronts. 
On one hand, neural and diffusion-based stylization~\citep{gatys2016,styleid,instantstyle,stylecrafter}, including hybrid pipelines that delegate appearance to generative rendering~\citep{coder,video_coder}, approximate style implicitly within a pretrained latent prior, making outputs prone to prior drift, stochastic inconsistency across long-form scenes, and difficulty in rule-level editing. 
On the other hand, visual-feedback refinement~\citep{manimtrainer,manimcode} and prompt-level stylization skills for coding agents either confine improvements to a single scene script or force agents to revisit references and reimplement visual details from scratch for every new scene. 
This raises a fundamental question: \textit{\textbf{Can artistic style be distilled from reference artworks into a reusable, transferable, and improvable executable program?}}

Artistic style is expressed through key drawing features such as color allocation and brushwork, which can be implemented through computational procedures. 
Classical painterly rendering has demonstrated the potential of such procedures to produce expressive artistic effects~\citep{hertzmann1998painterly}. 
Inspired by this insight, we introduce \textit{\textbf{CodeArtist}}, a Python-based painting engine that separates style-specific drawing capabilities from scene content and motion.
We decompose artistic styles into three complementary components: palette, brushwork, and texture. 
Palette captures the characteristic color distribution and relationships, brushwork describes the patterns and properties of strokes, and texture represents the surface qualities of different artistic media. 
Together, these components form an executable style representation that can be applied to diverse visual content while preserving its underlying artistic features.

Representing artistic style as modular, executable code yields three key advantages over both diffusion-based style transfer and agent-based stylization skills. 
First, style programs are \textbf{reusable}.
Once distilled, a style implementation can be invoked across new subjects, compositions, and timelines without revisiting the original reference images, ensuring deterministic visual consistency. 
Second, they are \textbf{transferable} across tasks and models.
By separating style development from downstream scene generation, mature drawing expertise built by frontier models can be directly inherited by lower-cost models through a common interface. 
Third, they are \textbf{improvable}.
A visible flaw in color allocation, contour construction, or stroke texture can be traced to a specific module and surgically revised in code, allowing a single improvement to accumulate and automatically benefit all downstream creations.
We conducted experiments across six coding agents from three model families, comparing \textit{CodeArtist} with agents that recreate each scene directly from reference artworks. 
The main results show that \textit{CodeArtist} improves visual quality across all models, with particular benefits for lower-cost models, while also reducing generation time and cost. 
Furthermore, we evaluate \textit{CodeArtist} in practical downstream tasks, including long-form video generation, coherent icon-set creation, cross-model refinement, targeted editing, and fast style adaptation, demonstrating its consistency, flexibility, and controllability across diverse creative workflows.
Overall, our contributions are threefold:
\begin{itemize}
\item We introduce \textit{CodeArtist}, a Python-based painting engine that represents artistic styles as modular executable programs by decoupling palette, brushwork, and texture from scene content and motion.
\item We show that executable style programs can be reused across scenes, transferred between different coding agents, and continuously improved through module-level updates.
\item Experiments across six coding agents demonstrate that \textit{CodeArtist} improves visual quality and style consistency while reducing generation time and cost across diverse creative tasks.
\end{itemize}

\section{Related Work}

\paragraph{Painterly and Stroke-Based Rendering.}
Modeling human artistic craft through explicit strokes and procedures has a long tradition in computer graphics. 
Classical painterly rendering translates photographs into hand-painted aesthetics via ordered, multi-scale brush strokes guided by image gradients and explicit rendering rules~\citep{haeberli1990paint,hertzmann1998painterly}.
More recent neural painting methods learn or optimize parametric brush strokes and vector curves using model-based reinforcement learning~\citep{huang2019learning}, optimal-transport stroke optimization~\citep{zou2021stylized}, feed-forward Transformer prediction~\citep{liu2021paint}, or CLIP-guided Bézier synthesis~\citep{frans2022clipdraw}, while graphics program induction recovers structured drawing code from simple hand-drawn inputs~\citep{ellis2018}.
However, these methods either solely reproduce a given input image into strokes, or synthesize new drawings with fixed naive primitives that cannot realize diverse, media-specific artistic styles.
In contrast, \textit{CodeArtist} uses coding agents to distill painterly traits, such as palette, brushwork, and texture, from reference artworks into modular, cross-media Python programs, enabling reusable, targetedly editable synthesis of novel scenes and animations.

\paragraph{Generative Style Transfer.}
Neural style transfer pioneered recombining content and artistic appearance through learned convolutional features~\citep{gatys2016}. 
More recently, diffusion-based methods achieve expressive image and video stylization via training-free attention injection~\citep{styleid}, decoupled feature adapters~\citep{instantstyle,stylecrafter}, and temporally constrained video translation~\citep{rerender}. 
However, because generative models approximate style implicitly within pretrained latent priors, they remain susceptible to prior drift, content-style entanglement, and stochastic inconsistency over long horizons, and cannot be inspected or surgically edited at the rule level.
Taking a different path, \textit{CodeArtist} encodes style as transparent, deterministic drawing programs, enforcing explicit stylistic rules free from prior drift and further guaranteeing cross-scene consistency.

\paragraph{Code-Driven Visual Generation.}
Coding agents are increasingly used to produce illustrations, educational graphics, and general-purpose animations as executable scripts, supported by benchmarks evaluating end-to-end delivery~\citep{code2video2025chen,vode2videocompass2026chen,vibevideobench2026diao}.
To improve visual quality, recent works apply rendering feedback and visual self-correction to refine scene code~\citep{manimtrainer,manimcode}, or equip coding agents with specialized drawing, plotting, and animation skills in Python~\citep{vonhahn2026matplotlib,chien2026manim,cloveric2026drawing}.
However, these approaches remain tied to individual scene scripts or fixed style presets: visual improvements do not carry over to new scenes, and artistic styles cannot be systematically distilled from reference artworks and transferred across tasks and models.
Drawing inspiration from reusable program and skill libraries~\citep{dreamcoder2021ellis,codeaspolicies2023liang,lilo2024grand,voyager2023wang}, \textit{CodeArtist} elevates artistic style itself into persistent, executable programs, allowing module-level refinements and upgrades to accumulate across scenes and enabling lower-cost models to inherit mature drawing capabilities built by frontier models.

\section{Method}
\begin{figure}
    \centering
    \includegraphics[width=0.99\linewidth]{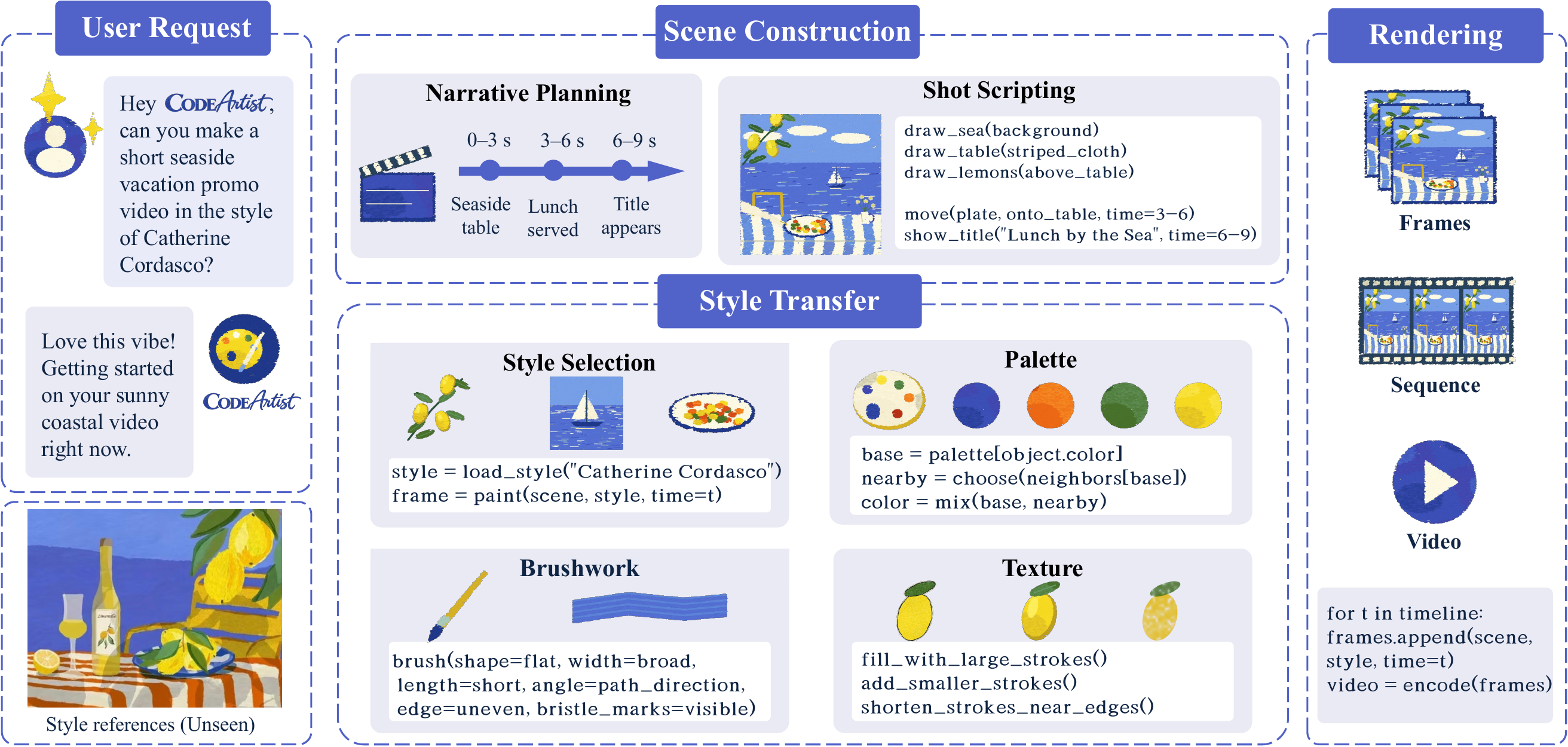}
    
    \caption{Overview of the \textit{CodeArtist} pipeline. \textit{CodeArtist} converts a natural-language request into a narrative timeline and shot scripts, applies the target palette, brushwork, and texture during frame generation, and encodes the resulting sequence as a stylized video.}
    \label{fig:main}
\end{figure}

\subsection{Structure}
 \paragraph{Overall workflow.}
As shown in Figure~\ref{fig:main}, \textit{CodeArtist} performs artistic style transfer through two stages: scene construction and style rendering. 
In the scene construction stage, a coding agent interprets the natural-language request to determine the target content, artistic style, and timing, and writes executable scene code. 
For videos, the narrative is first divided into a timeline of contiguous \textit{shots}, where each shot is a short scene segment implemented by a Python script specifying its layout and motion over local time. 
Static images follow the same workflow as a single-frame special case. 
Rather than painting pixels directly, running the active shot script at timestamp $t$ produces a style-agnostic list of drawing primitives, such as polygons, strokes, and text, each carrying an object identity and a semantic color role. 
Because this intermediate representation is independent of style, the same scene can be re-rendered in different artistic styles without modifying its content or motion code. 
In the style rendering stage, the \textit{CodeArtist} engine executes the selected style program on these primitives, applying its palette, brushwork, and texture directly as each element is drawn rather than adding a style filter afterward. 

\paragraph{Style representation.}
We represent each artistic style using three complementary components: palette, brushwork, and texture, each capturing a distinct aspect of its visual characteristics.
\textbf{Palette} describes the color distribution and statistics in reference artworks, including the proportions of different colors and their roles in the composition. 
It assigns style-specific colors to visual elements such as backgrounds, outlines, main objects, and highlights.
\textbf{Brushwork} determines how strokes are rendered along a given path. 
Brush marks are generated procedurally by sampling points along the path, with their orientation, width, and spacing controlled by local direction and pressure. 
\textbf{Texture} defines the surface appearance associated with different artistic media. 
For watercolor, it simulates transparent pigment layers, irregular wet edges, and color accumulation on paper. 
For oil painting, it combines layered underpainting with overlapping bristle strokes to create thick paint and visible brush traces.
These three components work together to translate an artistic style into executable rendering rules, allowing the same visual content to be rendered in different artistic styles.

\subsection{Evaluation}
We use a vision-language model (VLM) as a judge to evaluate style transfer across six artistic dimensions. 
For each task, we derive a task-specific rubric from expert analyses of the reference artwork, translating its characteristic visual features into concrete evaluation criteria. 
Given the reference artwork, generated image, and task requirements, the judge scores each dimension independently.
Specifically, $S_{\mathrm{series}}$ assesses overall series coherence, $S_{\mathrm{composition}}$ evaluates compositional structure, and $S_{\mathrm{color}}$ examines color and lighting consistency. 
Meanwhile, $S_{\mathrm{material}}$ verifies authentic medium fidelity, $S_{\mathrm{abstraction}}$ measures alignment in the level of abstraction, and $S_{\mathrm{completion}}$ ensures uniform stylization across all scene elements.
Each dimension is scored on a scale from 0 to 10, with higher scores indicating better alignment with the reference style.
Since the primary objective is to determine whether the generated image belongs to the same visual series as the reference, \(S_{\mathrm{series}}\) receives 50\% of the total weight, while the remaining dimensions contribute 10\% respectively.
These combine into the Visual Score $V$ metric:
\begin{equation}
    V=5 \cdot S_{\mathrm{series}} + S_{\mathrm{composition}} + S_{\mathrm{color}} + S_{\mathrm{material}} + S_{\mathrm{abstraction}} +  S_{\mathrm{completion}}
\end{equation}

\section{Experiments}

\paragraph{Experimental setup.}
We evaluate six models from three families: Claude-Opus-5.5 and Claude-Haiku-5.5~\citep{anthropic2026opus55,anthropic2026haiku55}, GPT-6-Astra and GPT-5.6-Sol~\citep{openai2026gpt6astra,openai2026gpt56sol}, and DeepSeek-V4.1-Flash and DeepSeek-V4-Flash~\citep{deepseek2026v41,deepseek2026v4}. 
For each model family, we select a frontier model and a less capable counterpart to explore whether CodeArtist’s drawing capabilities can be shared across models with different capabilities.
We use the low reasoning-effort setting for all runs during evaluation. 
DeepSeek and Claude models run through Claude Code, while GPT models run through Codex. 
Each model is evaluated in Harbor under two code-generation settings. 

\textbf{Baseline.}
As our baseline, the model receives the reference artwork and implements the requested scene from scratch. 
In contrast, \textit{CodeArtist} uses an existing style program to implement the content, but does not see the reference artwork. 
This setting tests whether artistic style represented with code can replace repeated visual imitation and transfer across models and tasks.

\textbf{Metric.}
We evaluate all methods on 30 tasks using four metrics. 
\textbf{Visual Score} is assessed by Gemini-3.8-Flash using the six-dimensional rubric described above. 
\textbf{Time} and \textbf{Cost} report the average execution time and cost per task.
\textbf{Agreement} measures the consistency between VLM and human pairwise preferences.
Specifically, let $h_i, a_i \in \{-1, 0, 1\}$ denote human and VLM preferences for pair $i$, where $1$ favors \textit{CodeArtist}, $-1$ favors the baseline, and $0$ indicates a tie. 
Agreement is defined as the fraction of non-tied pairs for which the human and VLM preferences match:
$\mathrm{Agreement}=\frac{|\mathcal{M}|}{|\mathcal{V}|}$
where $\mathcal{M}=\{i:h_i=a_i\neq 0\}$ and $\mathcal{V}=\{i:h_i\neq 0,\ a_i\neq 0\}$.

\subsection{Results}

Table~\ref{tab:main_results} compares \textit{CodeArtist} with the baseline across six models from three model families. \textit{CodeArtist} improves the visual score for every model, increasing the average from 34.93 to 38.42. 
It also reduces average runtime from $8.07$ to $4.10$ minutes and average cost from \$0.4036 to $\$0.2053$, corresponding to reductions of 49.2\% and 49.1\%, respectively. 
\textit{CodeArtist} benefits different model tiers in complementary ways.
Specifically, it improves the visual capabilities of lower-cost models while reducing the generation overhead of frontier models through reusable style implementations.
Human-evaluator preference agreement ranges from 65.43\% to 90.67\%, with a macro-average of 74.83\% across models.

\begin{figure}[t]
    \centering
    \begin{minipage}[c]{0.48\linewidth}
        \centering
        \includegraphics[width=\linewidth]{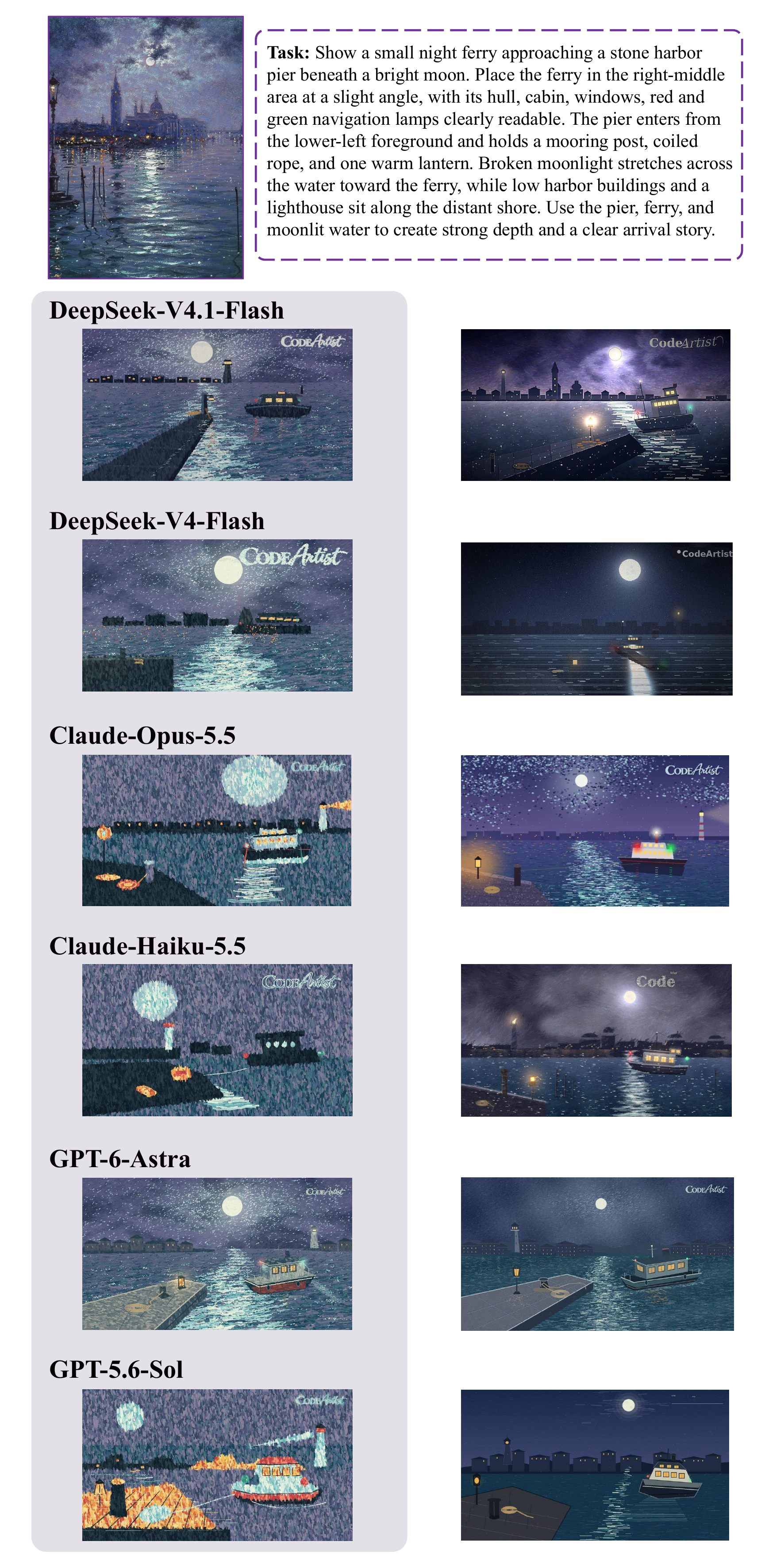}
        \caption{Qualitative comparison between CodeArtist (left) and the baseline (right) across six coding agents. 
        CodeArtist preserves more consistent palette, brushwork, and texture across model families.}
        \label{fig:moon}
    \end{minipage}
    \hspace{0.02\linewidth}
    \begin{minipage}[c]{0.48\linewidth}
        \centering
        \includegraphics[width=\linewidth]{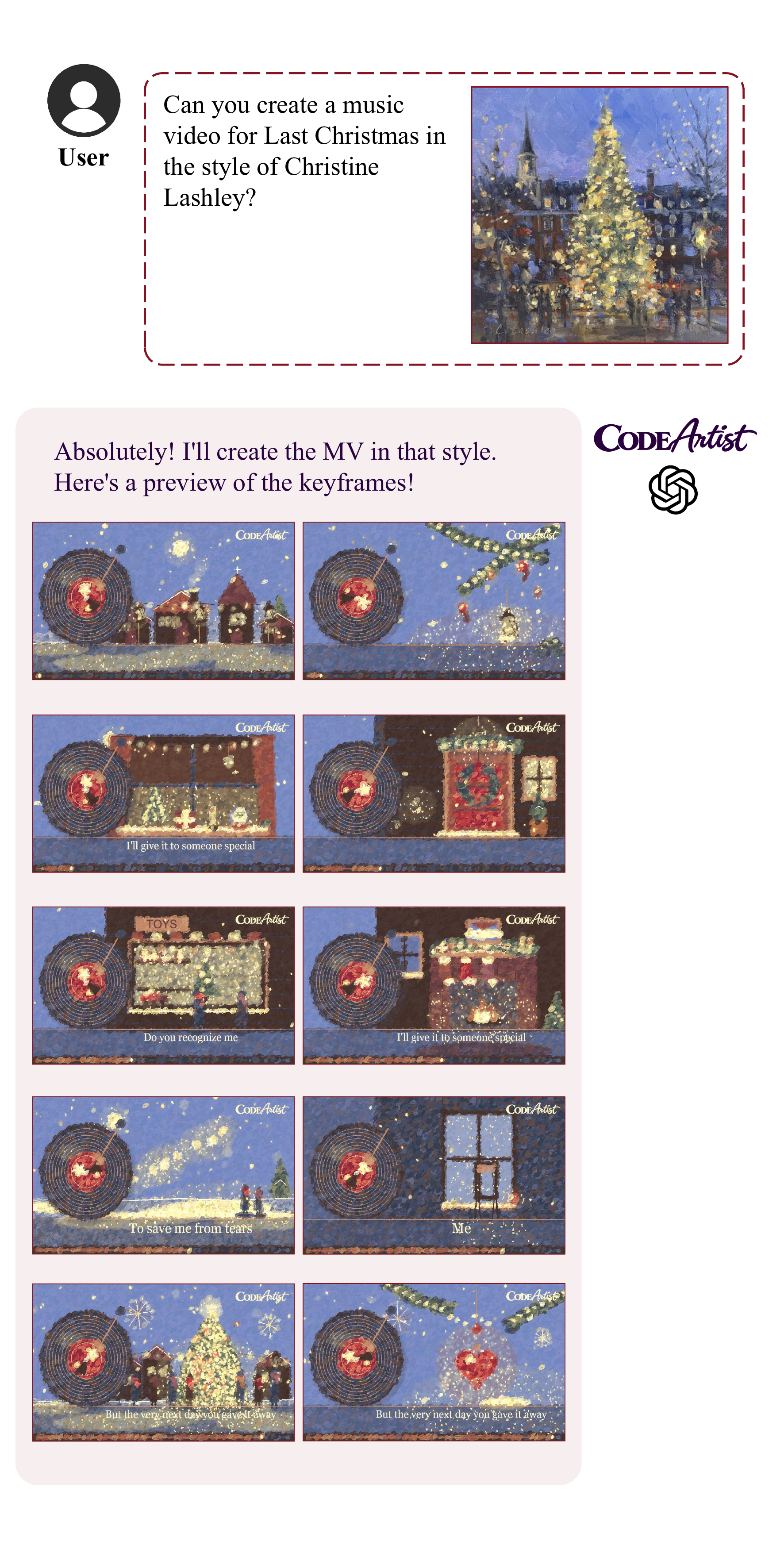}
        \caption{Long-form music video generated with CodeArtist. 
        Despite changes in scene content, composition, and lighting across shots, the frames retain a coherent visual style.}
        \label{fig:long}
    \end{minipage}
\end{figure}

As shown in Figure~\ref{fig:moon}, \textit{CodeArtist} expresses more stable style characteristics across model families.
Figures~\ref{fig:long} and~\ref{fig:icon} evaluate whether \textit{CodeArtist} remains stable in practical downstream tasks beyond single-image generation. 
Figure~\ref{fig:long} presents a complete music video in which the same programmatic style is reused across multiple shots with different subjects, layouts, and lighting conditions. 
Despite these scene changes, the video maintains a coherent palette, brushwork, and overall visual identity. 
Figure~\ref{fig:icon} shows a different use case of generating the mini-icons used in our main figure. 
Here, one style program is applied to objects with diverse shapes and semantic roles while keeping the icon set visually unified. Together, these examples show the potential of code in preserving a consistent artistic language across scenes, compositions, and visual asset types, making executable styles practical for real-world tasks.

\begin{table*}[t]
\centering
\caption{Overall comparison across model families. 
For each metric, we compare \textit{CodeArtist} against the baseline.
Best results in each pair are in \textbf{bold}.}
\label{tab:main_results}
\small
\setlength{\tabcolsep}{0pt}
\renewcommand{\arraystretch}{1.35}

\begin{tabular*}{\textwidth}{
@{\extracolsep{\fill}}
l
c c
c c
c c
c
@{}}
\toprule
\multirow{2}{*}{\textbf{Model}}
& \multicolumn{2}{c}{\textbf{Visual Score} $\uparrow$}
& \multicolumn{2}{c}{\textbf{Time (min)} $\downarrow$}
& \multicolumn{2}{c}{\textbf{Cost (\$)} $\downarrow$}
& \multirow{2}{*}{\textbf{Agreement} $\uparrow$} \\
\cmidrule(lr){2-3}
\cmidrule(lr){4-5}
\cmidrule(lr){6-7}
& \textbf{Ours} & Baseline
& \textbf{Ours} & Baseline
& \textbf{Ours} & Baseline
& \\
\midrule

\multicolumn{8}{l}{\textit{\textbf{DeepSeek}}} \\

\quad DeepSeek-V4-Flash
& \textbf{36.63} & 30.93
& \textbf{3.74} & 5.48
& \textbf{0.0171} & 0.0238
& 90.67\% \\

\quad DeepSeek-V4.1-Flash
& \textbf{41.37} & 37.13
& \textbf{2.94} & 9.44
& \textbf{0.0260} & 0.4072
& 70.51\% \\

\addlinespace[3pt]

\multicolumn{8}{l}{\textit{\textbf{GPT}}} \\

\quad GPT-5.6-Sol
& \textbf{36.30} & 33.40	
& \textbf{2.02} & 4.13
& \textbf{0.1873} & 0.2558
& 81.43\% \\

\quad GPT-6-Astra
& \textbf{41.83} & 38.63
& \textbf{9.00} & 16.33
& \textbf{0.7439} & 1.1543
& 66.67\% \\

\addlinespace[3pt]

\multicolumn{8}{l}{\textit{\textbf{Claude}}} \\

\quad Claude-Haiku-5.5
& \textbf{36.93} & 33.17
& \textbf{4.31} & 8.35
& \textbf{0.0098} & 0.0176
& 74.29\% \\

\quad Claude-Opus-5.5
& \textbf{37.47} & 36.33
& \textbf{2.56} & 4.68
& \textbf{0.2475} & 0.5628
& 65.43\% \\

\midrule

\textbf{Average}
& \textbf{38.42} & 34.93
& \textbf{4.10} & 8.07
& \textbf{0.2053} & 0.4036
& 74.83\% \\

\bottomrule
\end{tabular*}
\end{table*}

\subsection{Modular Refinement}
\textit{CodeArtist} is an initial implementation instead of a mature rendering system. 
As its artistic capabilities are represented as modular code, the engine can continue to evolve with frontier models and accumulate improvements contributed by the community. 
A stronger model can improve a shared implementation of color, brushwork, or texture, and the updated capability can become available to other models in downstream projects. 

\begin{figure}[t]
    \centering
    \begin{minipage}[c]{0.48\linewidth}
        \centering
        \includegraphics[width=\linewidth]{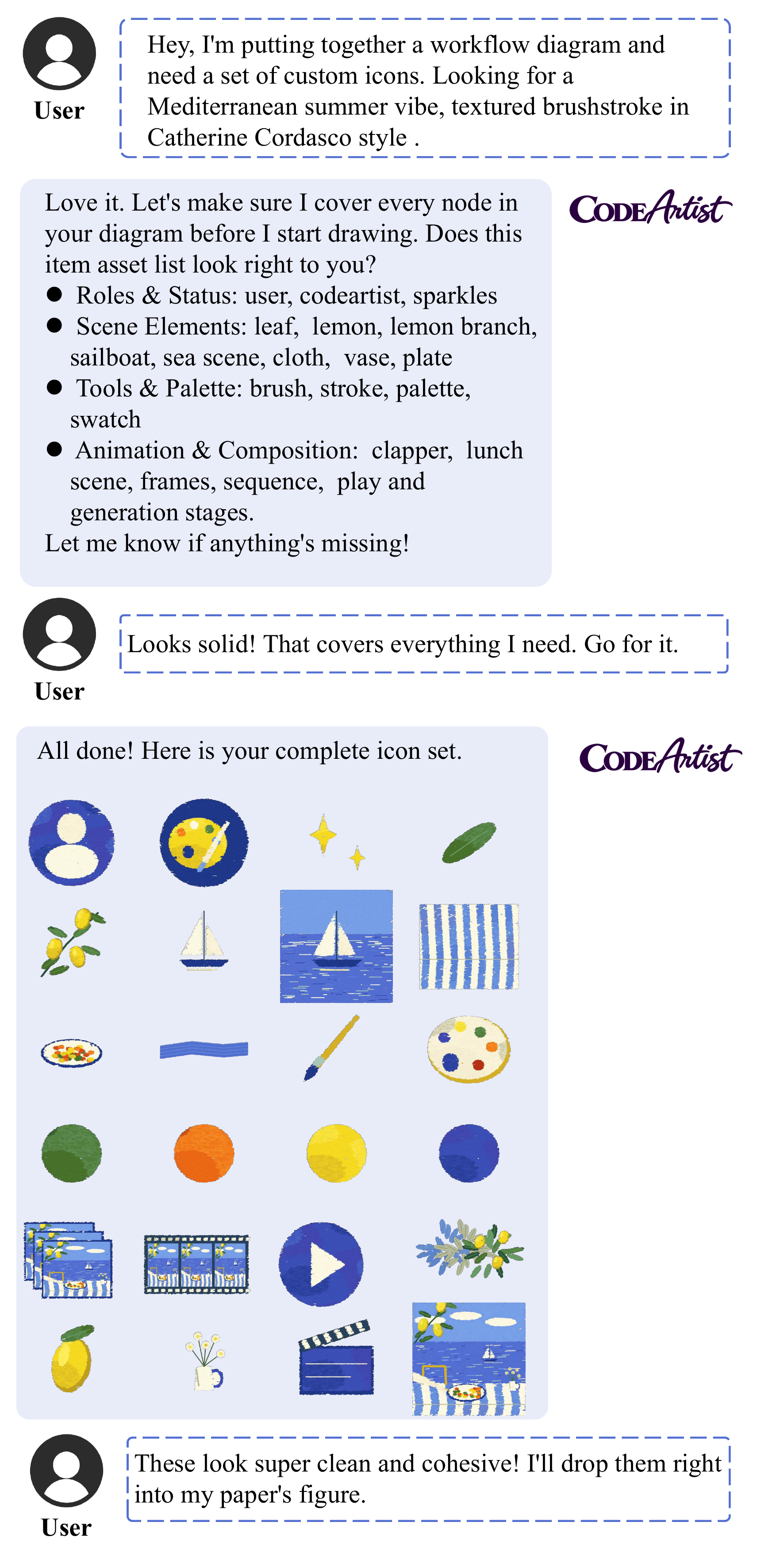}
        \caption{
        Icon set generated with a targeted style. 
        CodeArtist produces consistent colors and brushwork across diverse elements, from people and natural objects to symbols, tools, and scenes.}
        \label{fig:icon}
    \end{minipage}
    \hspace{0.02\linewidth}
    \begin{minipage}[c]{0.48\linewidth}
        \centering
        \includegraphics[width=\linewidth]{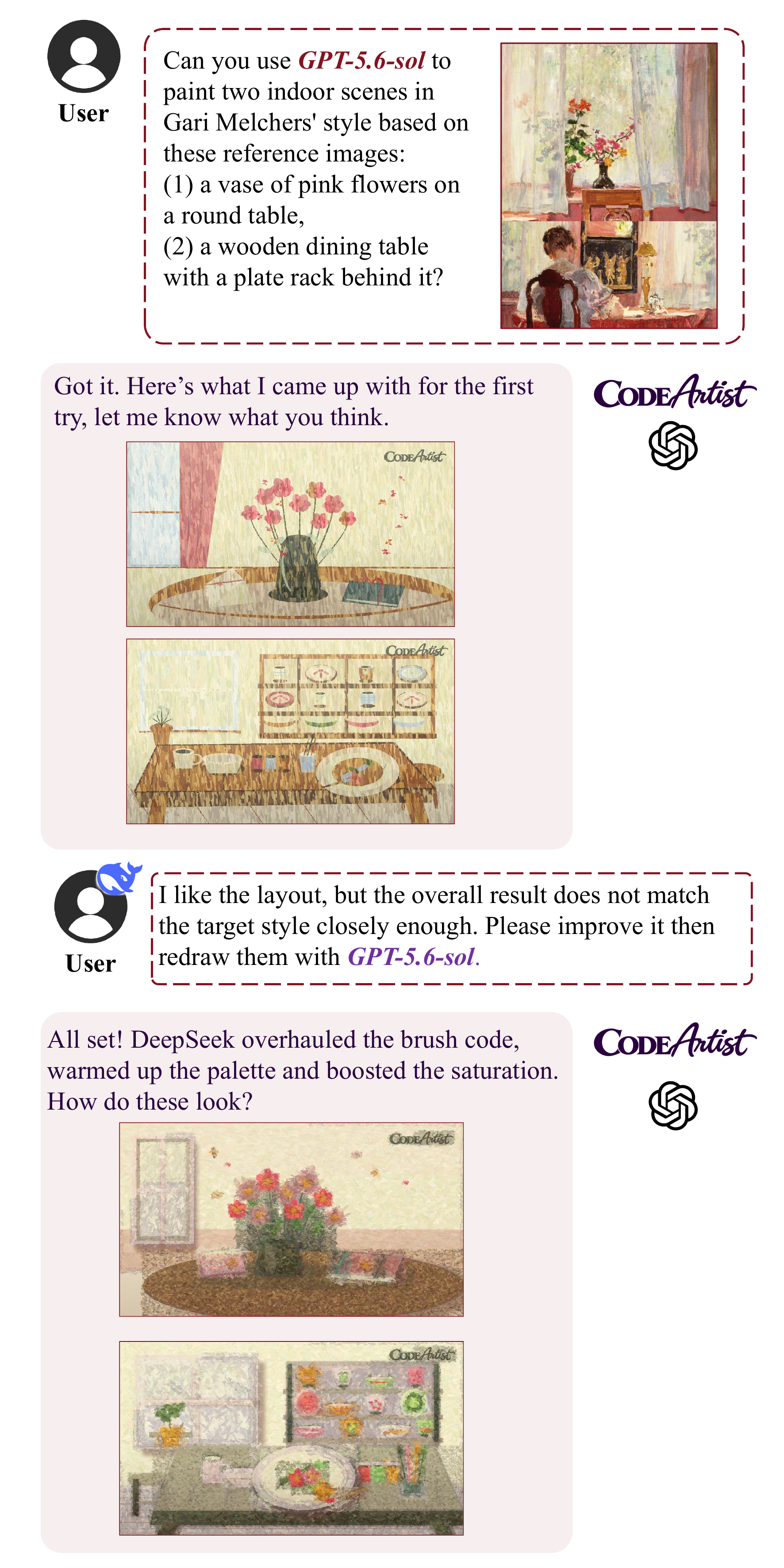}
        \caption{
        Cross-model refinement of a reusable style program. 
        GPT-5.6-Sol creates the initial scenes, DeepSeek-V4.1-Flash improves the shared functions, and GPT-5.6-Sol applies the updated modules to both scenes, enhancing texture, color warmth, and saturation.}
        \label{fig:refine}
    \end{minipage}
\end{figure}

Figure~\ref{fig:refine} demonstrates this process with a simple example. 
GPT-5.6-Sol first uses an existing style implementation to draw two indoor scenes. After the user notes that the initial results do not match the target style closely enough, DeepSeek-V4.1-Flash upgrades the shared brush and palette functions.
GPT-5.6-Sol then redraws the scenes using the revised implementation. 
The new results show stronger brush texture, warmer colors, and higher saturation, even though the final images are still generated by GPT-5.6-Sol. 
This example shows how improvements made by a more capable model can be preserved in the engine, shared across models, and reused by more practical projects.

\subsection{Targeted Modification}
\begin{figure}[t]
    \centering
    \begin{minipage}{0.45\linewidth}
        \centering
        \includegraphics[width=\linewidth]{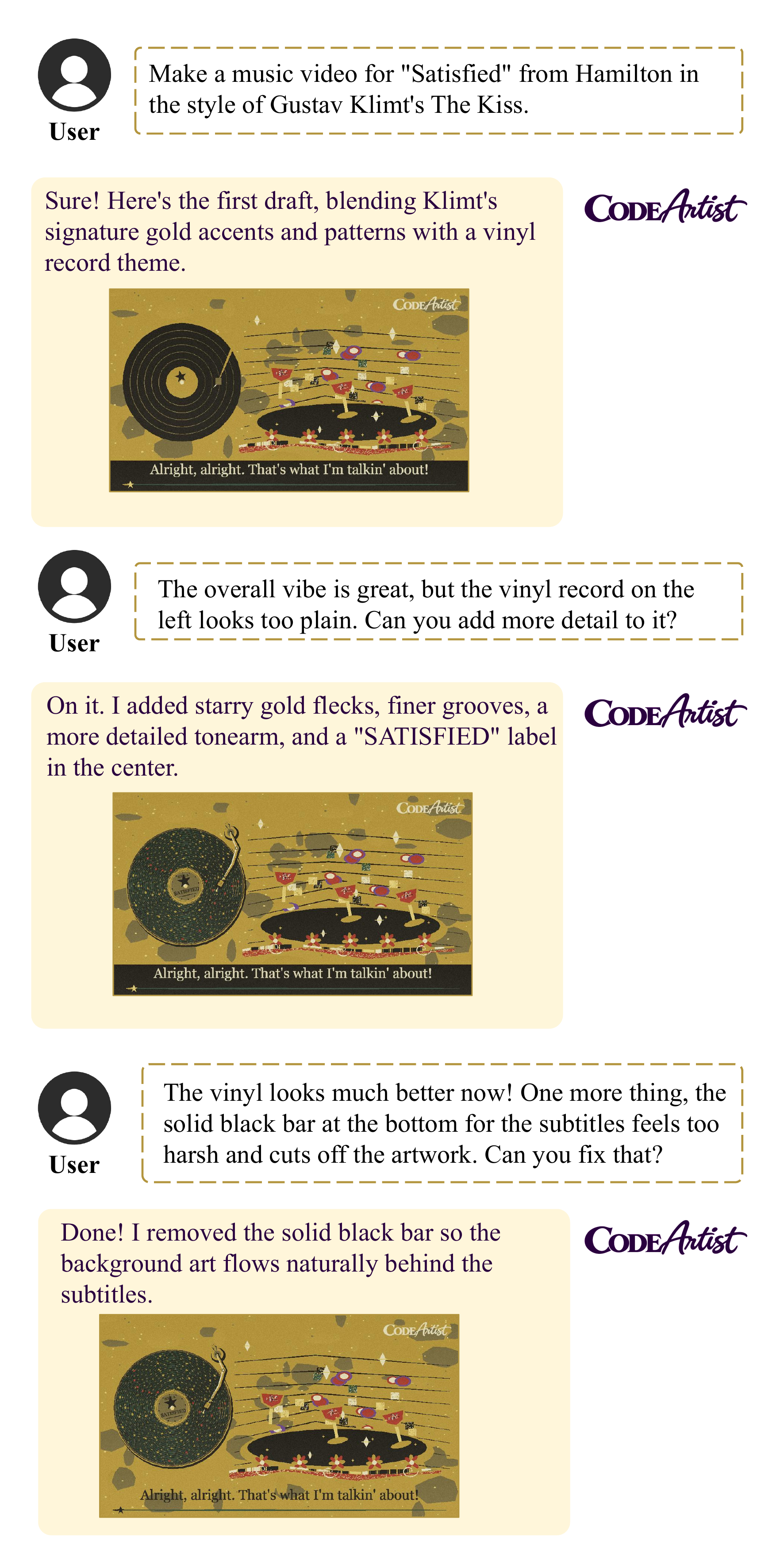}
        \caption{
        Targeted music-video editing. 
        \textit{CodeArtist} adds details to the vinyl record and removes the subtitle background while preserving the rest of the composition.}
        \label{fig:modi_ham}
    \end{minipage}\hfill
    \begin{minipage}{0.45\linewidth}
        \centering
        \includegraphics[width=\linewidth]{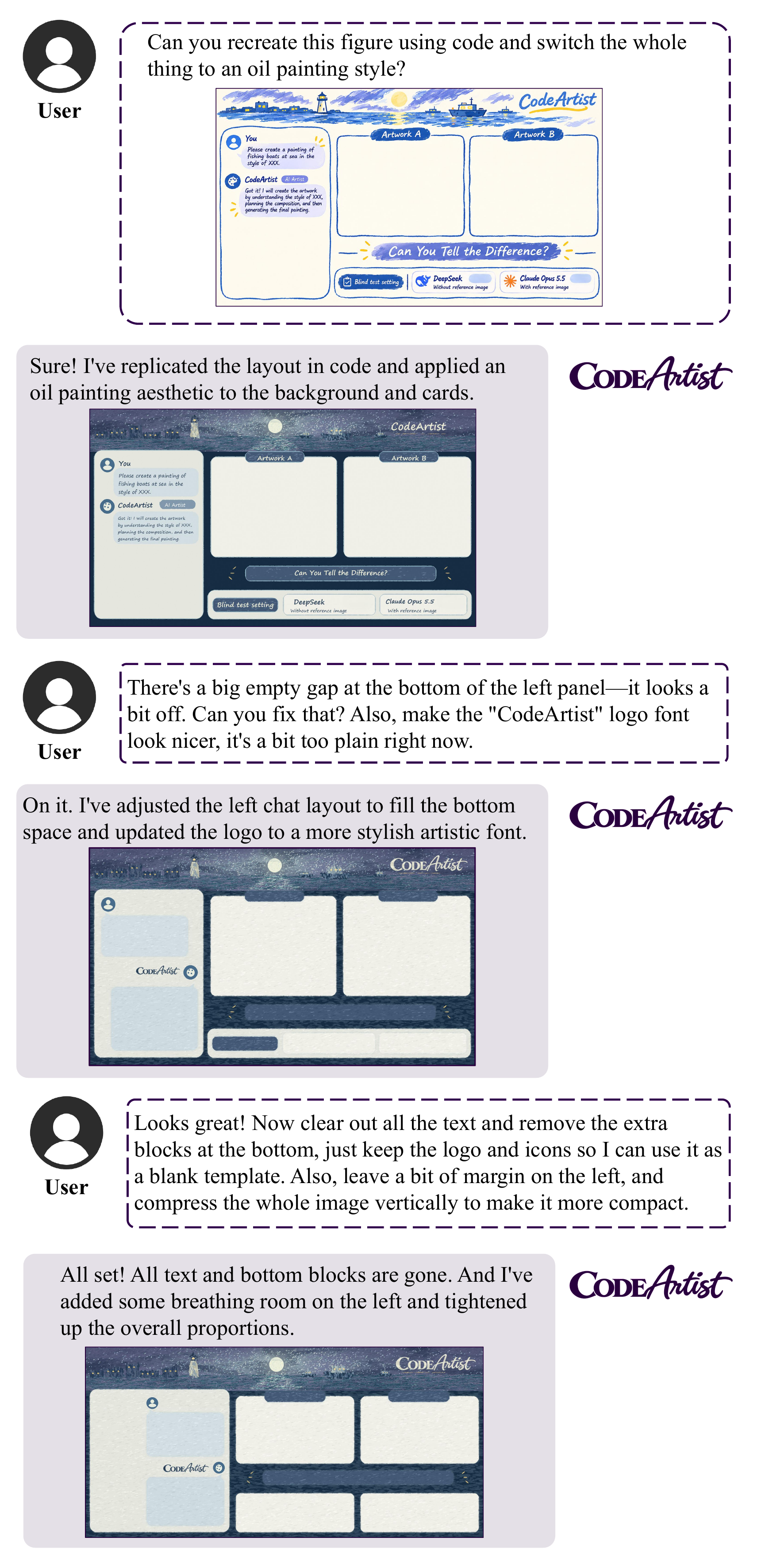}
        \caption{
        Iterative refinement of the teaser. \textit{CodeArtist} adjusts layout, typography, and content while preserving the overall oil-painting style.}
        \label{fig:modi_figure}
    \end{minipage}
\end{figure}

To assess \textit{CodeArtist}'s practicality in real-world user workflows, we evaluate whether it supports controllable, targeted revisions without disrupting unrelated elements.
Figure~\ref{fig:modi_ham} illustrates two localized edits to a music-video scene. \textit{CodeArtist} first adds a detailed tonearm to the vinyl record and then removes the solid subtitle bar, all while preserving the background and Klimt-inspired aesthetic. 
Figure~\ref{fig:modi_figure} further demonstrates this practical utility in multi-turn research figure editing, where the agent adjusts whitespace, removes specific text and interface blocks, and alters canvas proportions. 
Together, these cases show that content, layout, and style can be refined independently while keeping the unrelated components intact.

\subsection{Fast Style Adaptation}
\begin{figure*}[ht]
\centering
\setlength{\tabcolsep}{2pt}
\renewcommand{\arraystretch}{1.05}

\begin{tabular}{@{}ccc@{\hspace{8mm}}ccc@{}}

\includegraphics[width=.12\textwidth]{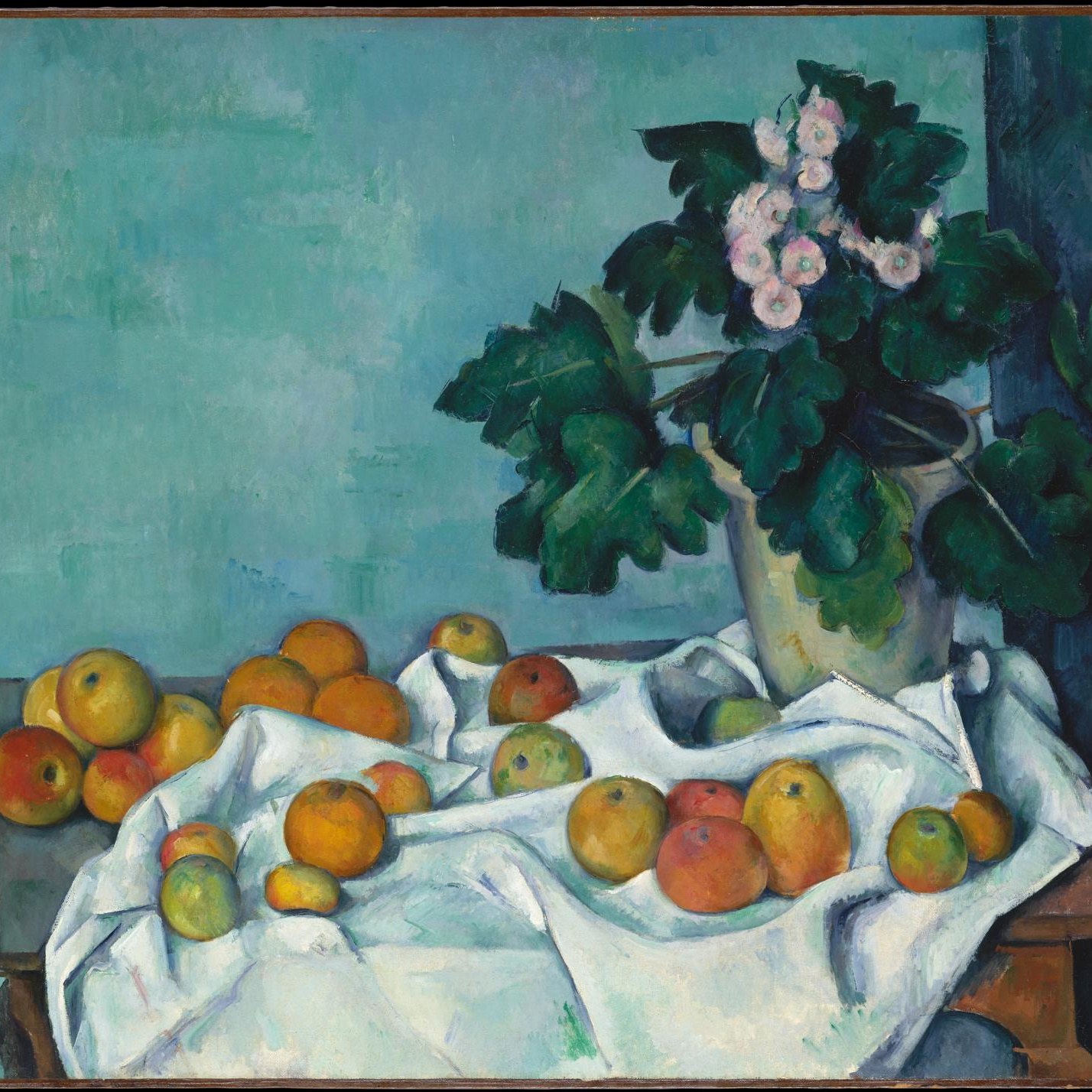} &
\includegraphics[width=.12\textwidth]{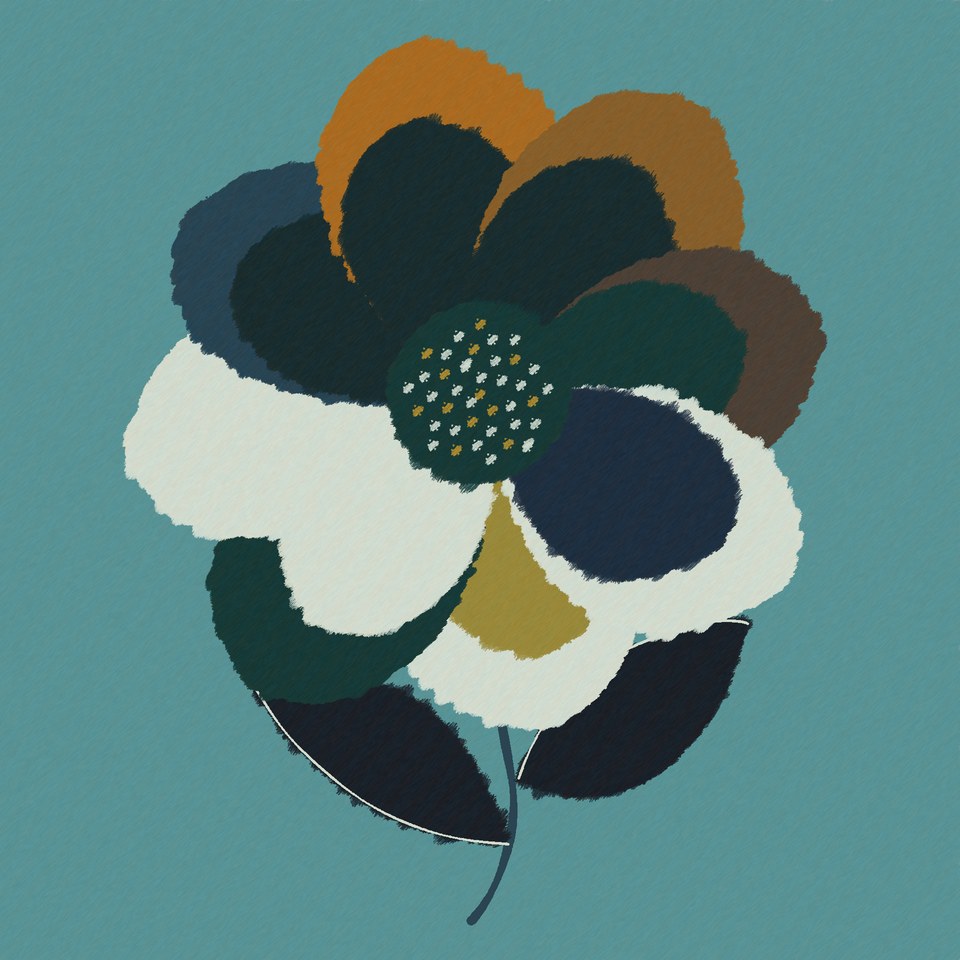} &
\includegraphics[width=.12\textwidth]{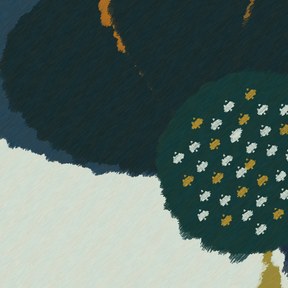} &
\includegraphics[width=.12\textwidth]{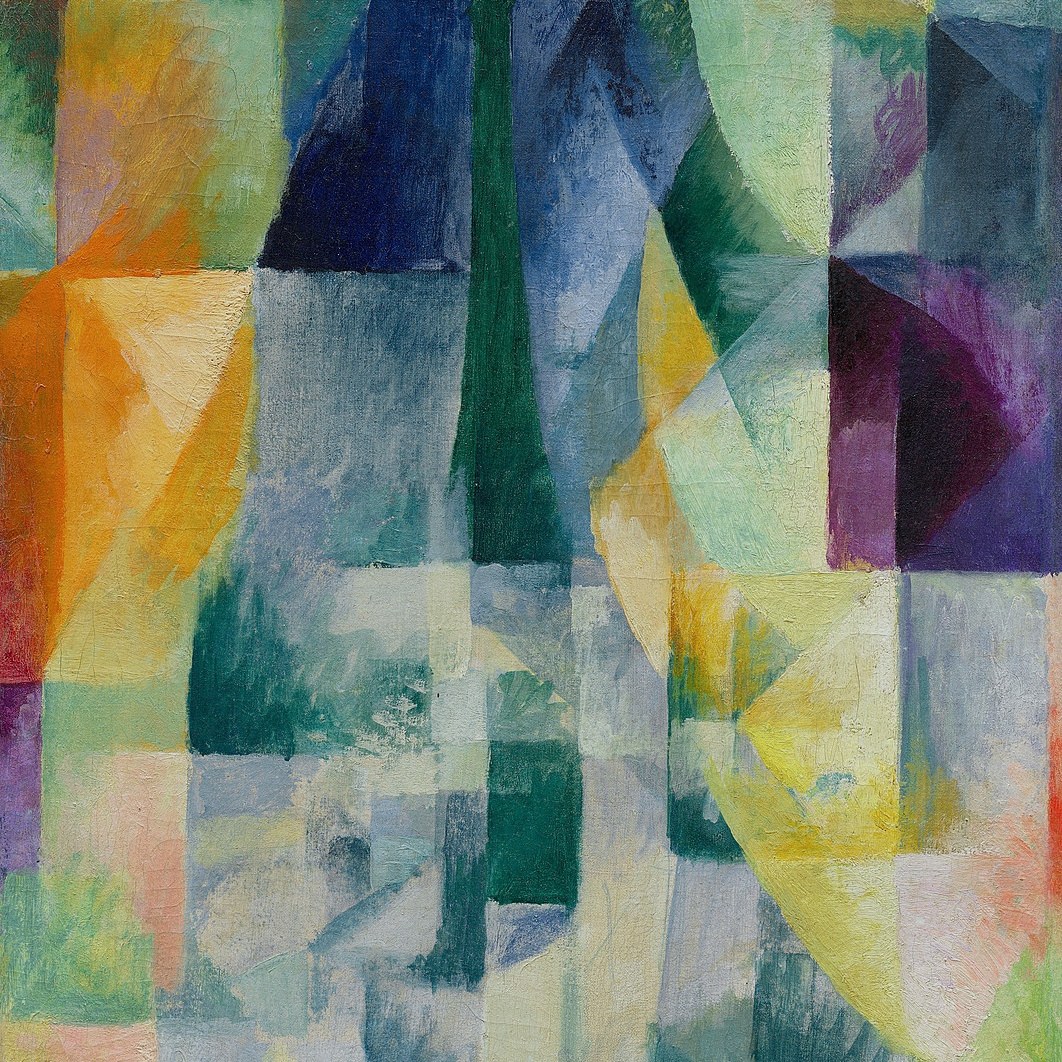} &
\includegraphics[width=.12\textwidth]{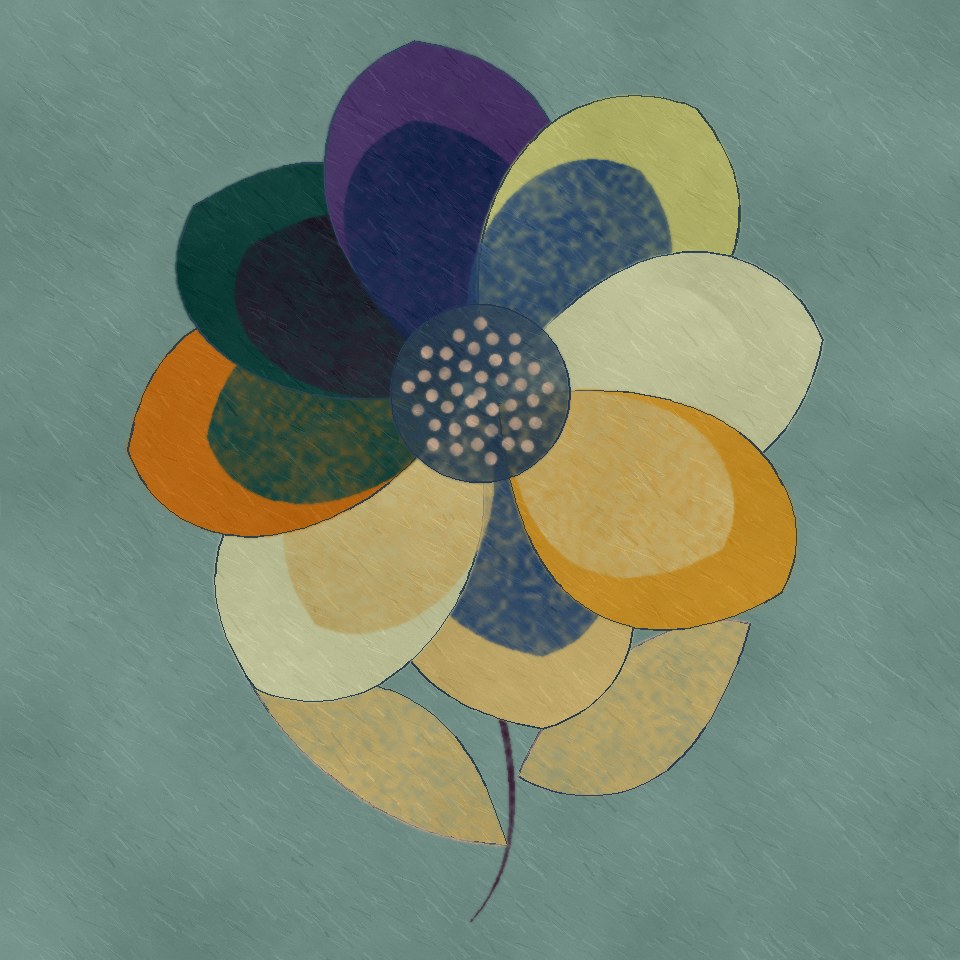} &
\includegraphics[width=.12\textwidth]{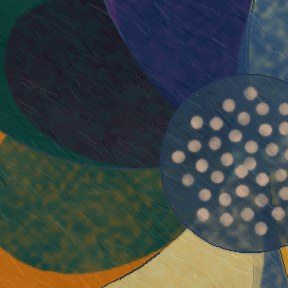}
\\[-1mm]
& \textsc{Cezanne} & & & \textsc{Delaunay} &
\\[3mm]

\includegraphics[width=.12\textwidth]{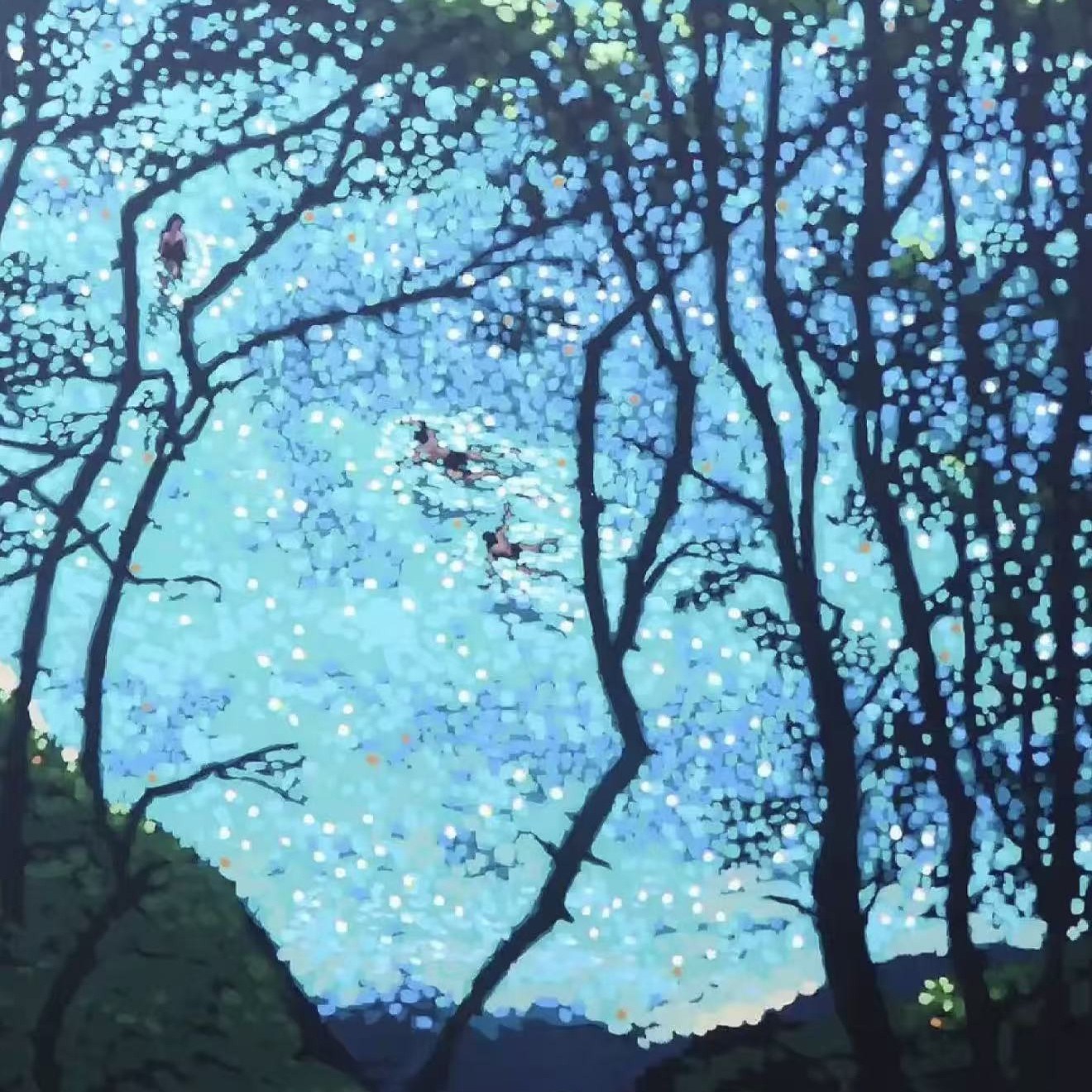} &
\includegraphics[width=.12\textwidth]{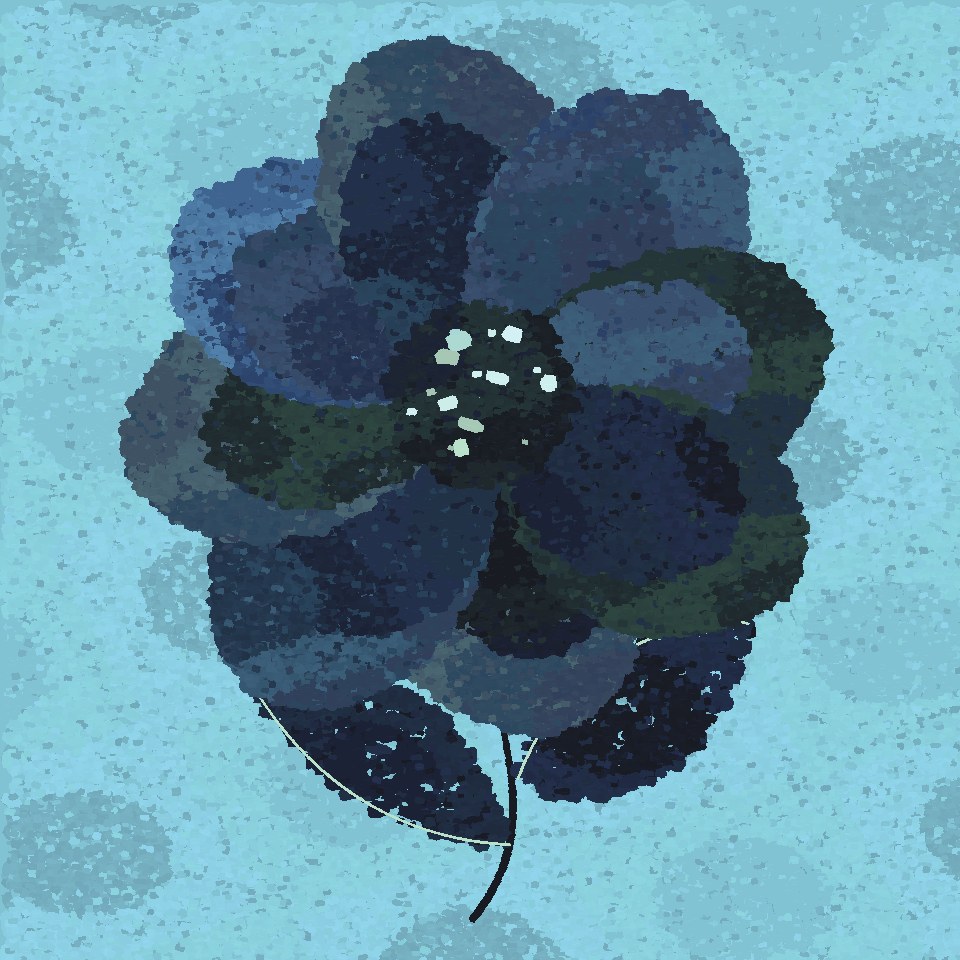} &
\includegraphics[width=.12\textwidth]{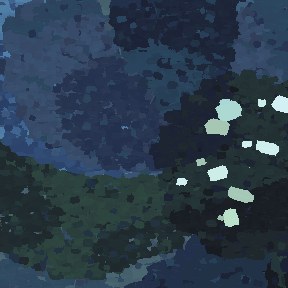} &
\includegraphics[width=.12\textwidth]{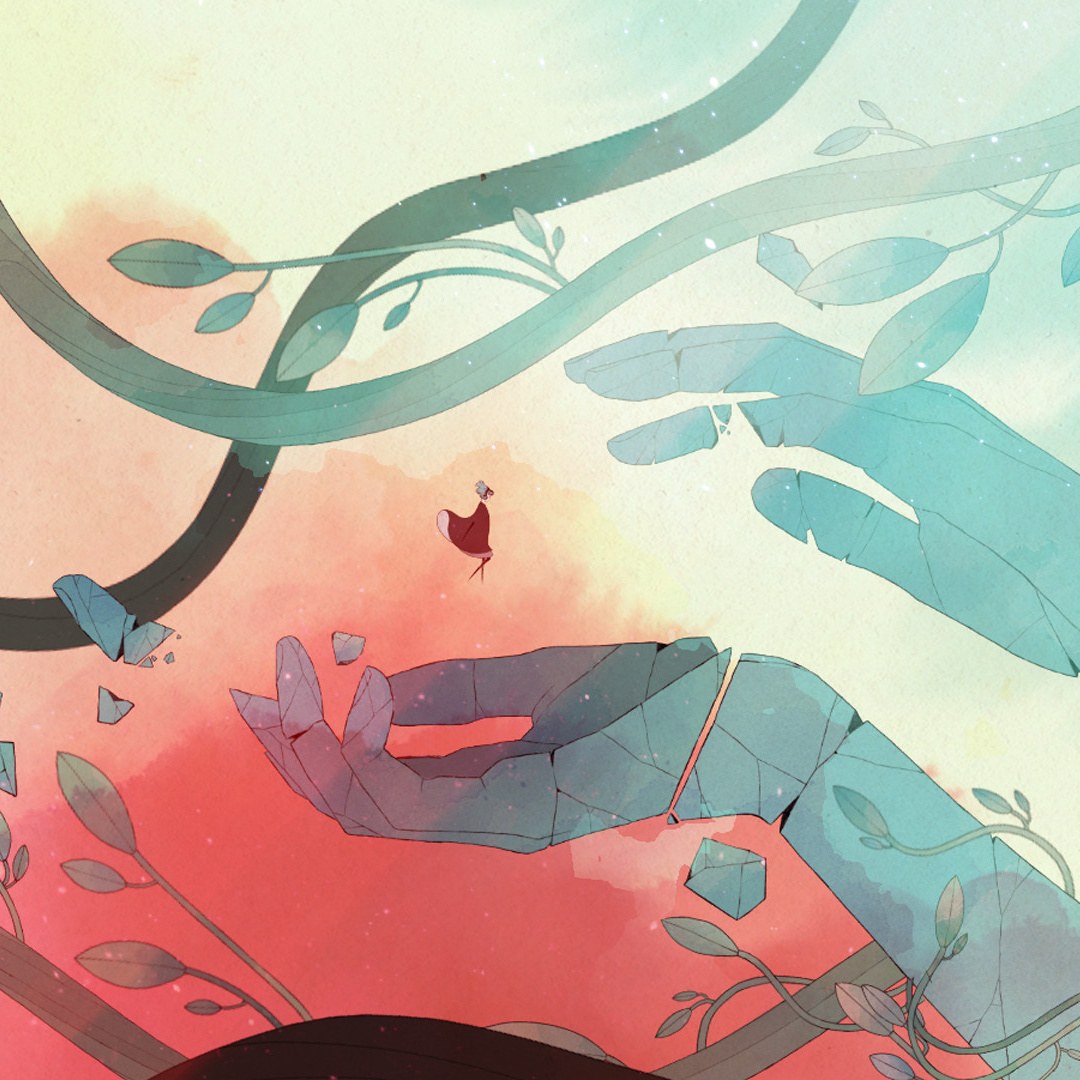} &
\includegraphics[width=.12\textwidth]{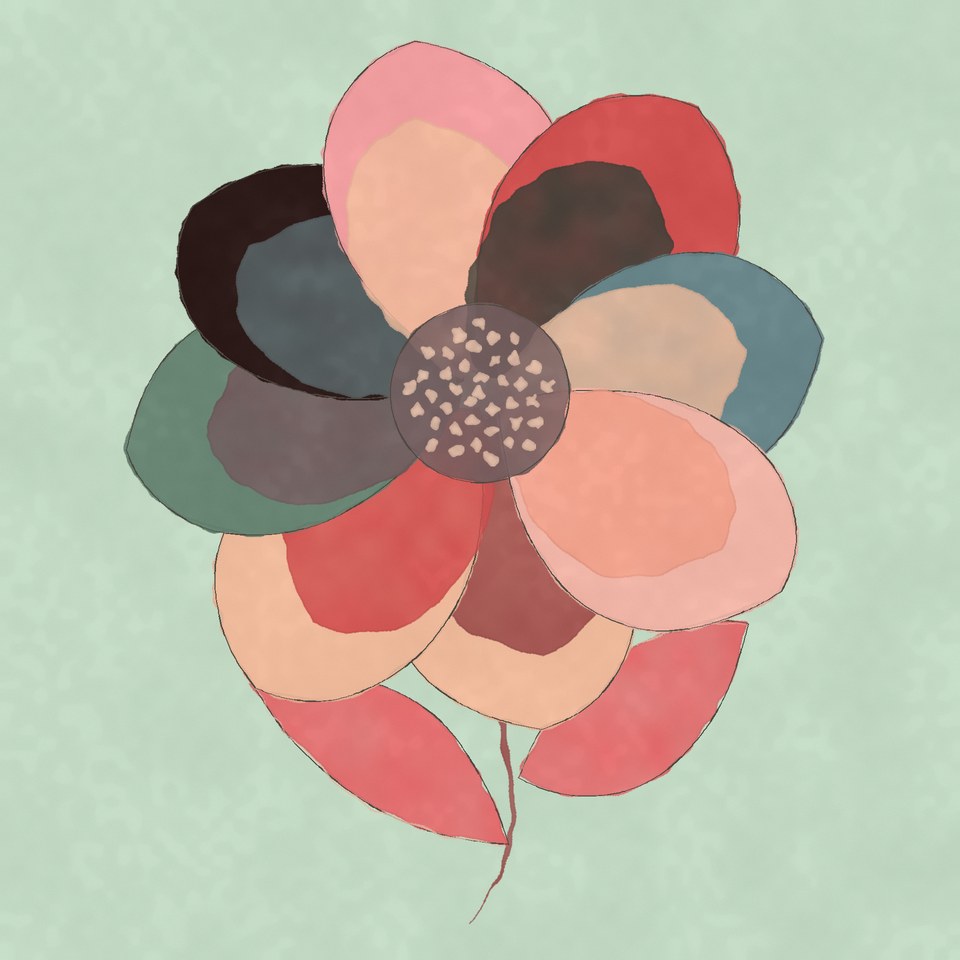} &
\includegraphics[width=.12\textwidth]{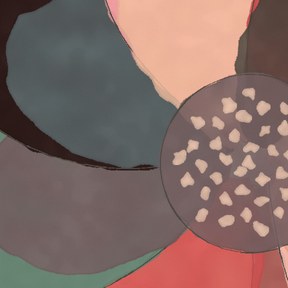}
\\[-1mm]
& \textsc{Gordon} & & & \textsc{Gris} &
\\[3mm]

\includegraphics[width=.12\textwidth]{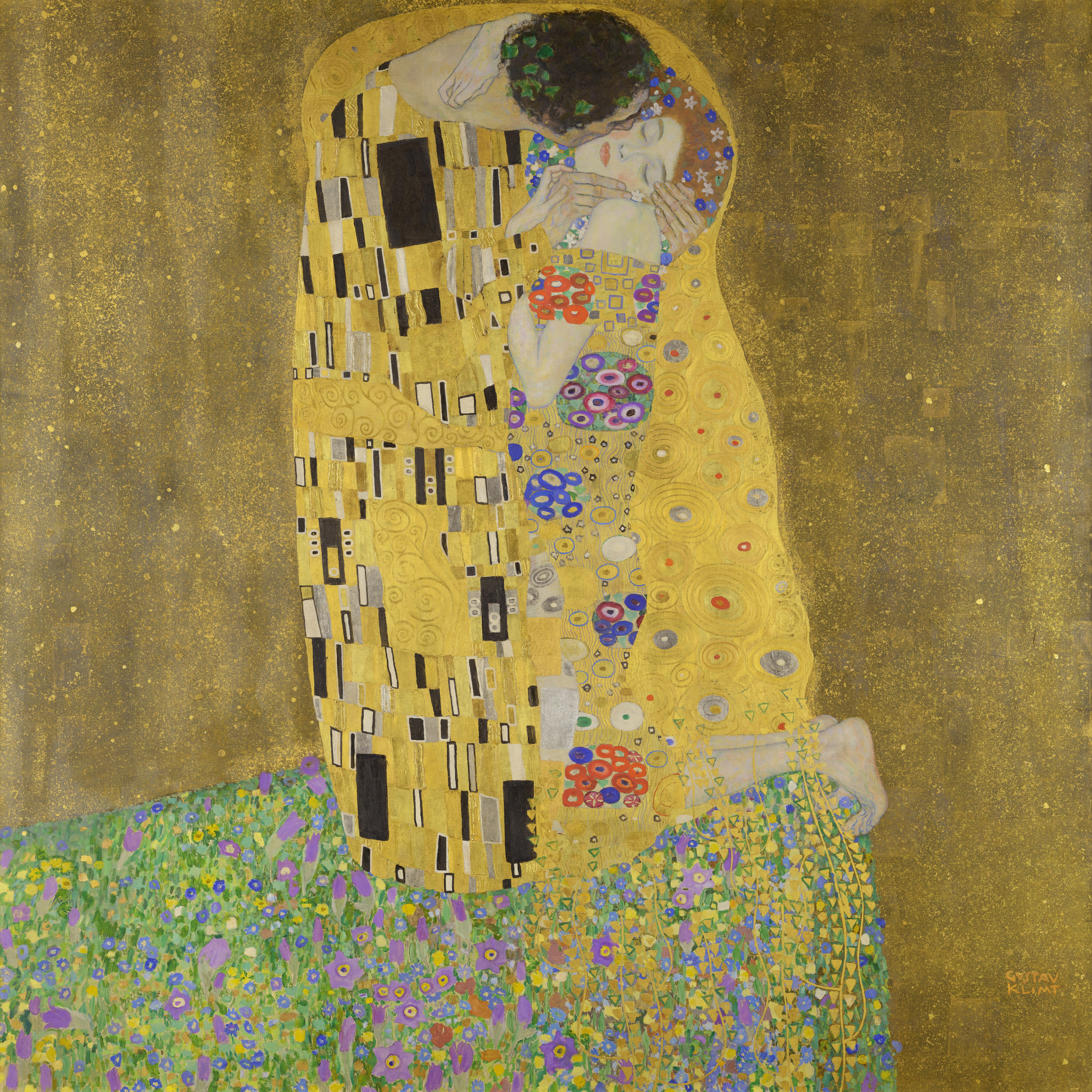} &
\includegraphics[width=.12\textwidth]{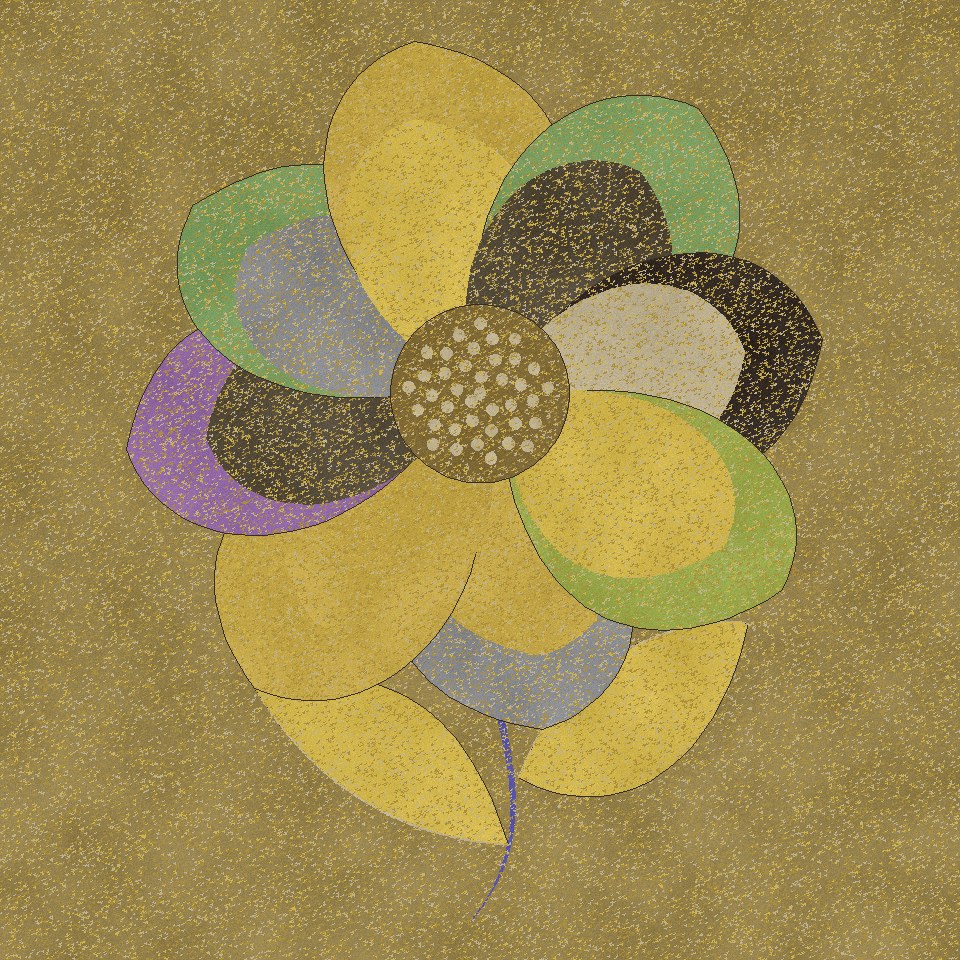} &
\includegraphics[width=.12\textwidth]{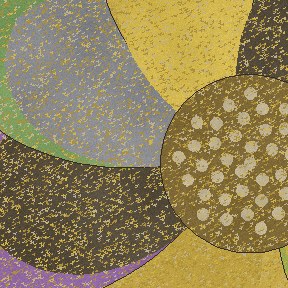} &
\includegraphics[width=.12\textwidth]{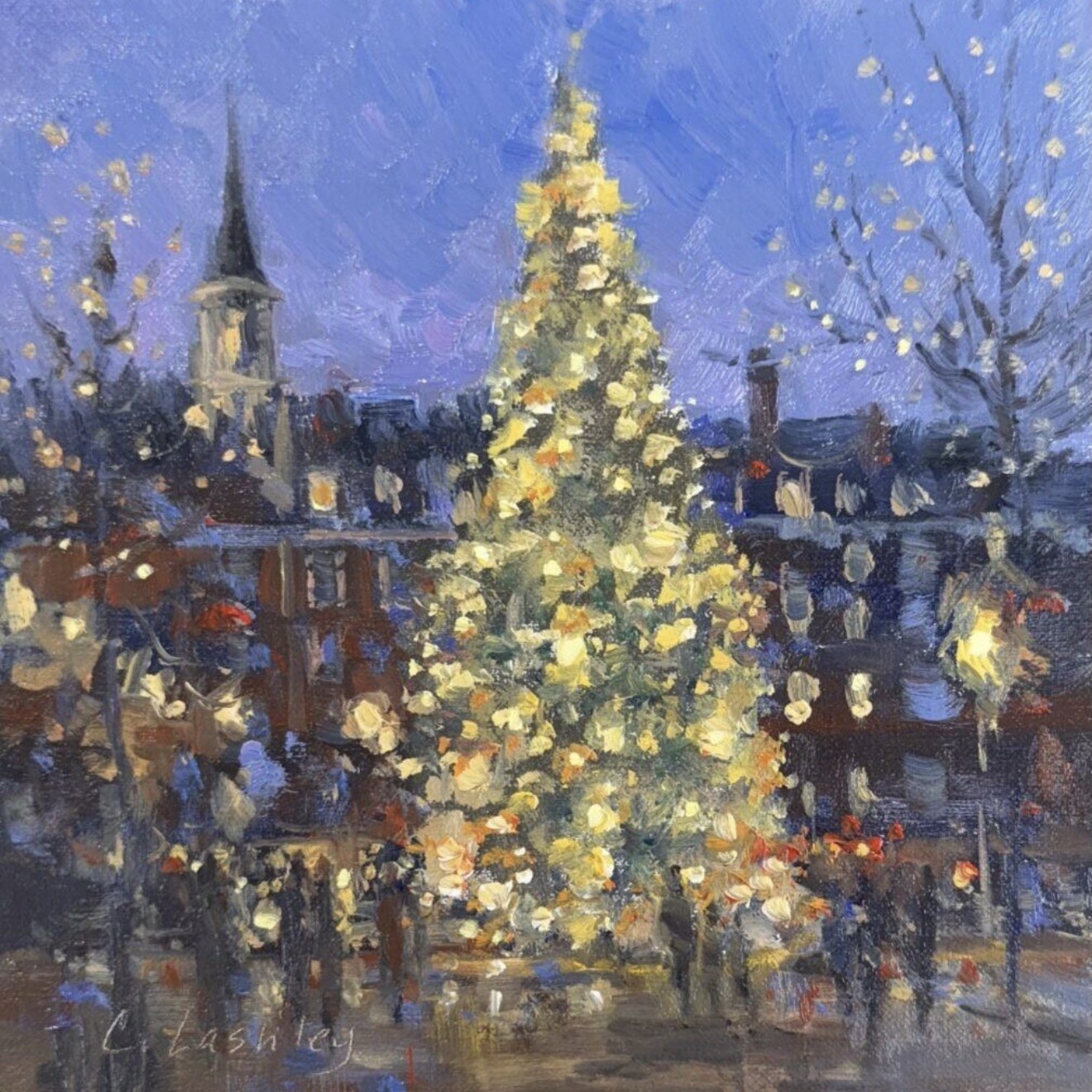} &
\includegraphics[width=.12\textwidth]{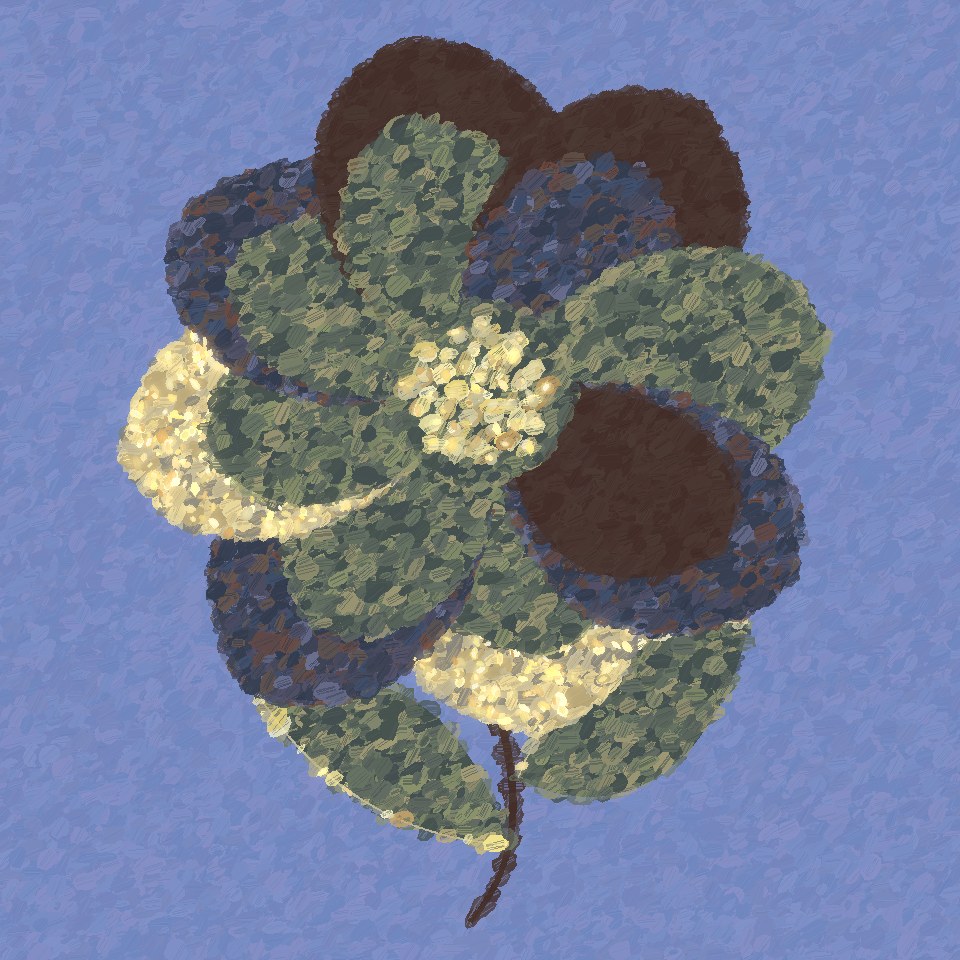} &
\includegraphics[width=.12\textwidth]{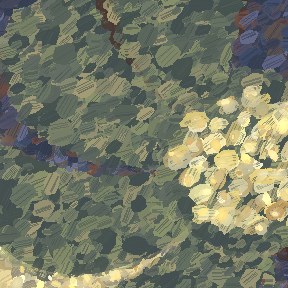}
\\[-1mm]
& \textsc{Klimt} & & & \textsc{Lashley} &
\\[3mm]

\includegraphics[width=.12\textwidth]{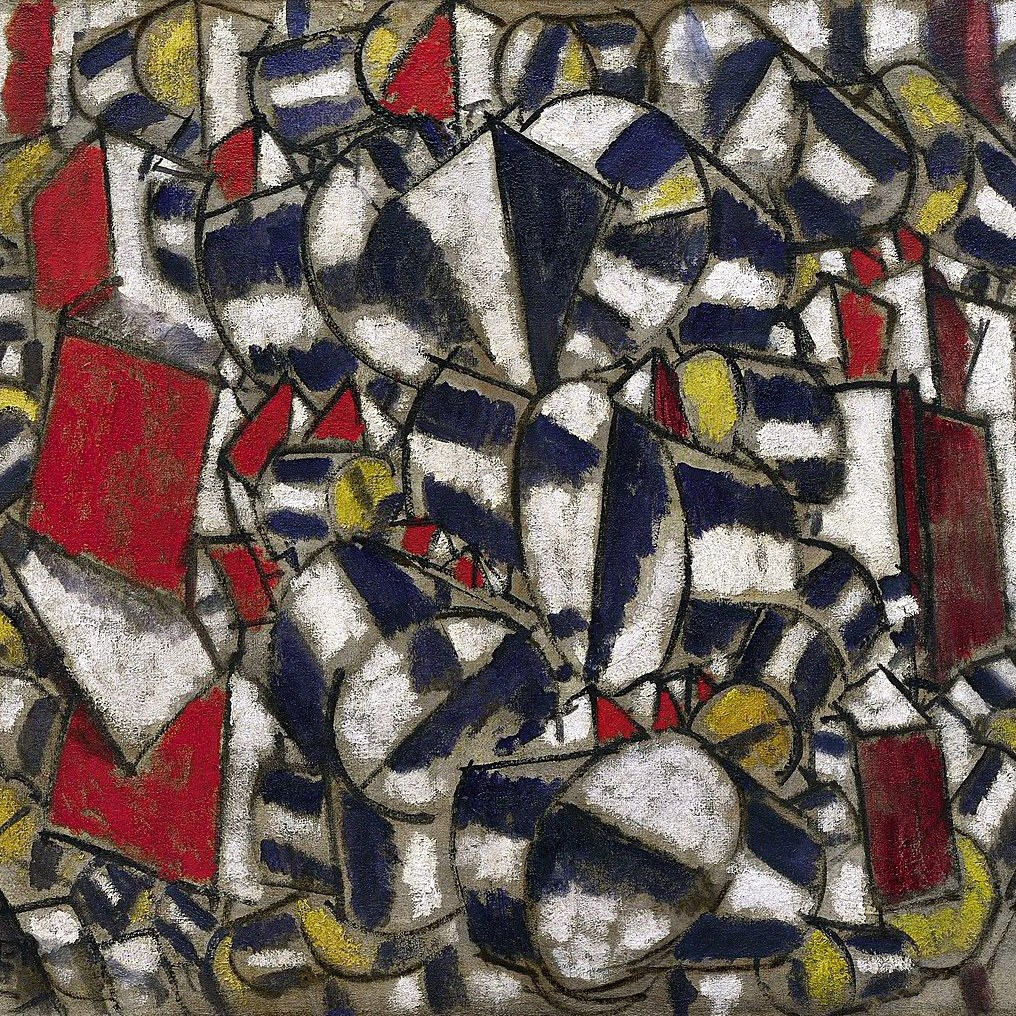} &
\includegraphics[width=.12\textwidth]{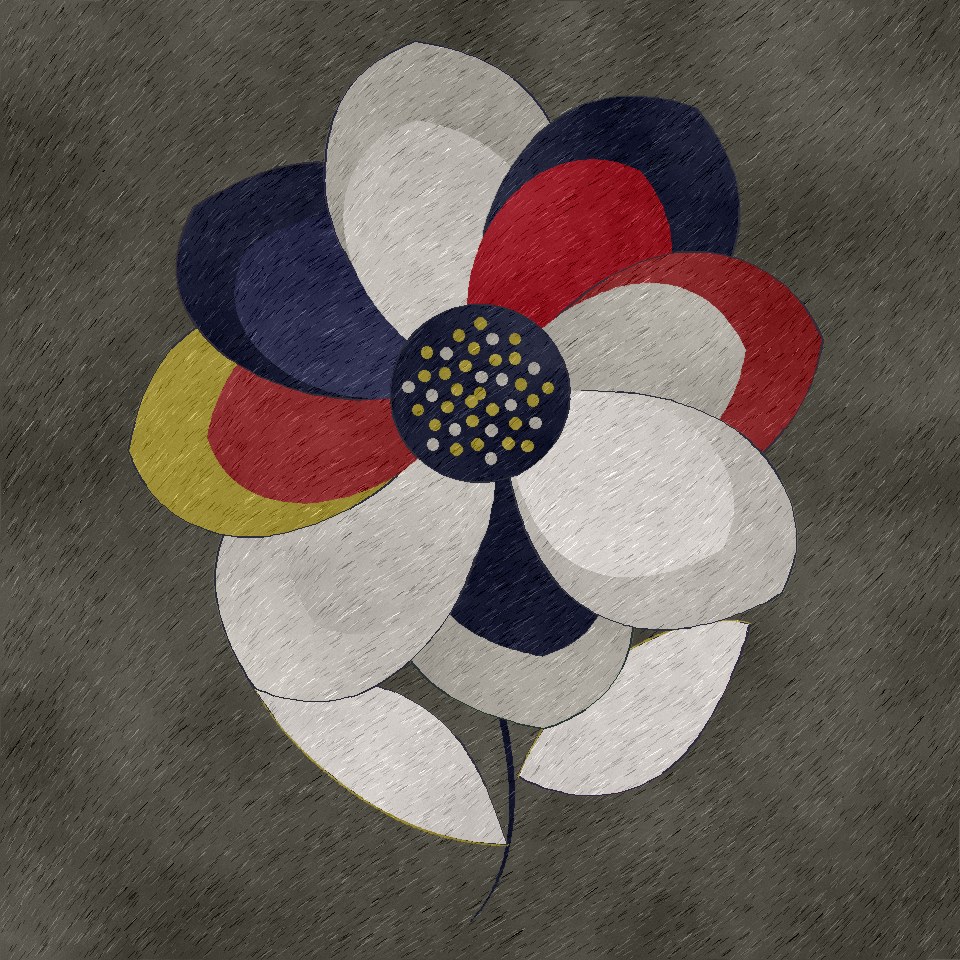} &
\includegraphics[width=.12\textwidth]{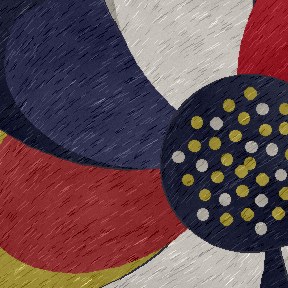} &
\includegraphics[width=.12\textwidth]{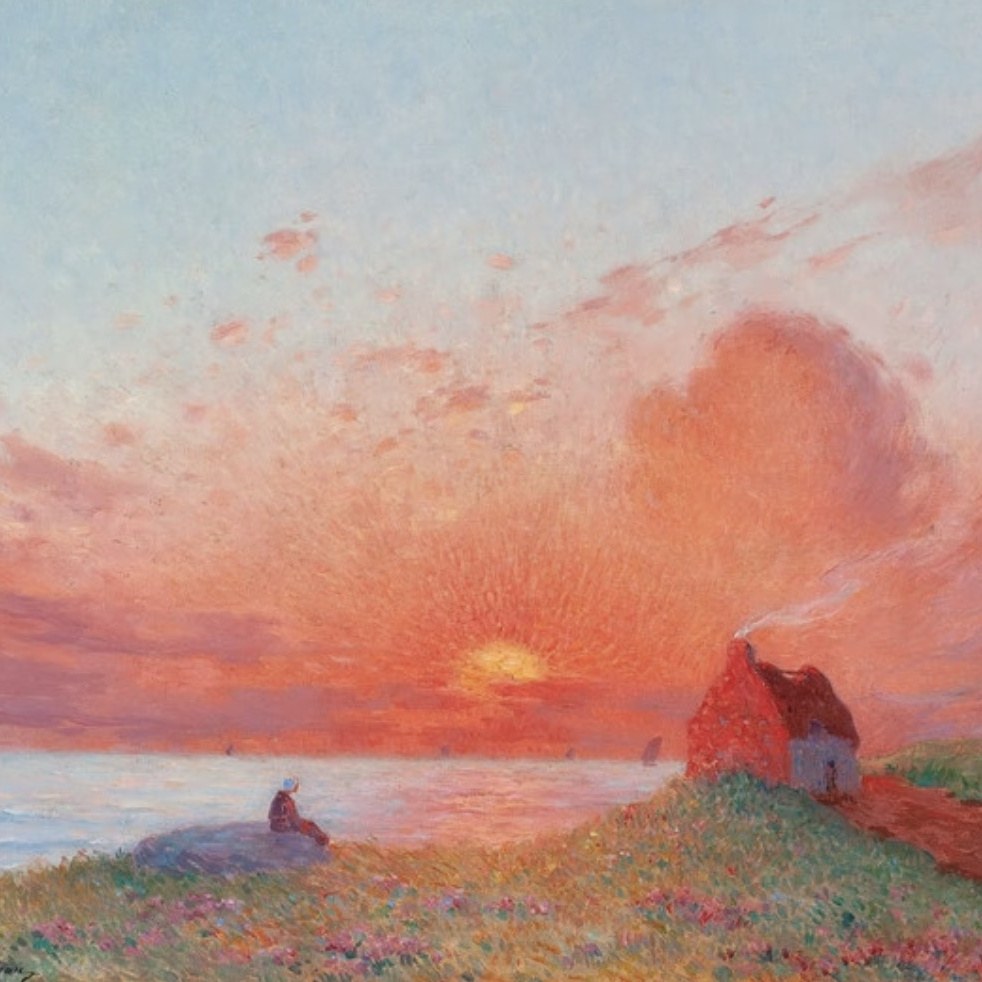} &
\includegraphics[width=.12\textwidth]{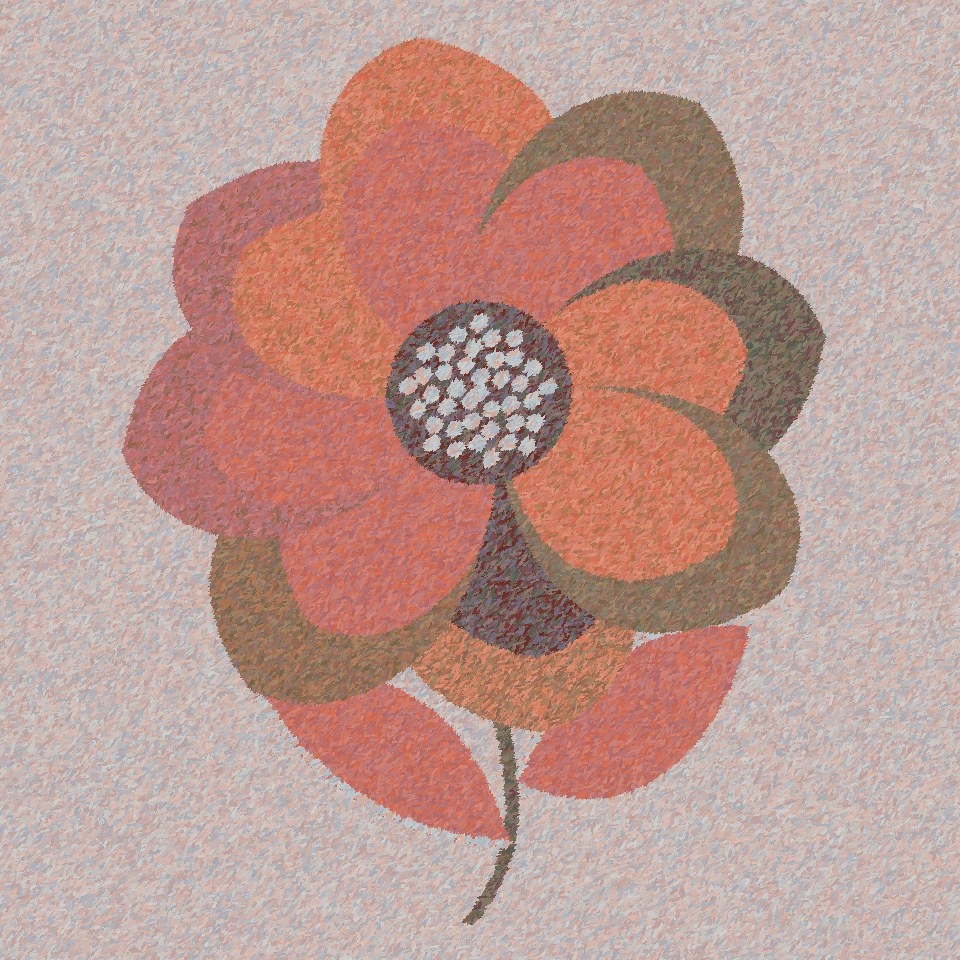} &
\includegraphics[width=.12\textwidth]{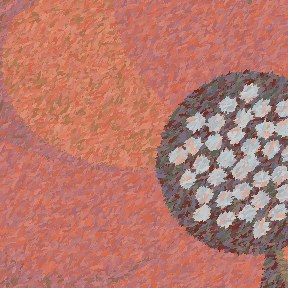}
\\[-1mm]
& \textsc{Leger} & & & \textsc{Lorn} &
\\[3mm]

\includegraphics[width=.12\textwidth]{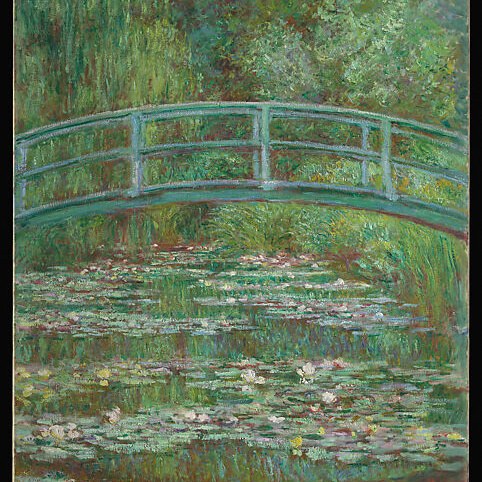} &
\includegraphics[width=.12\textwidth]{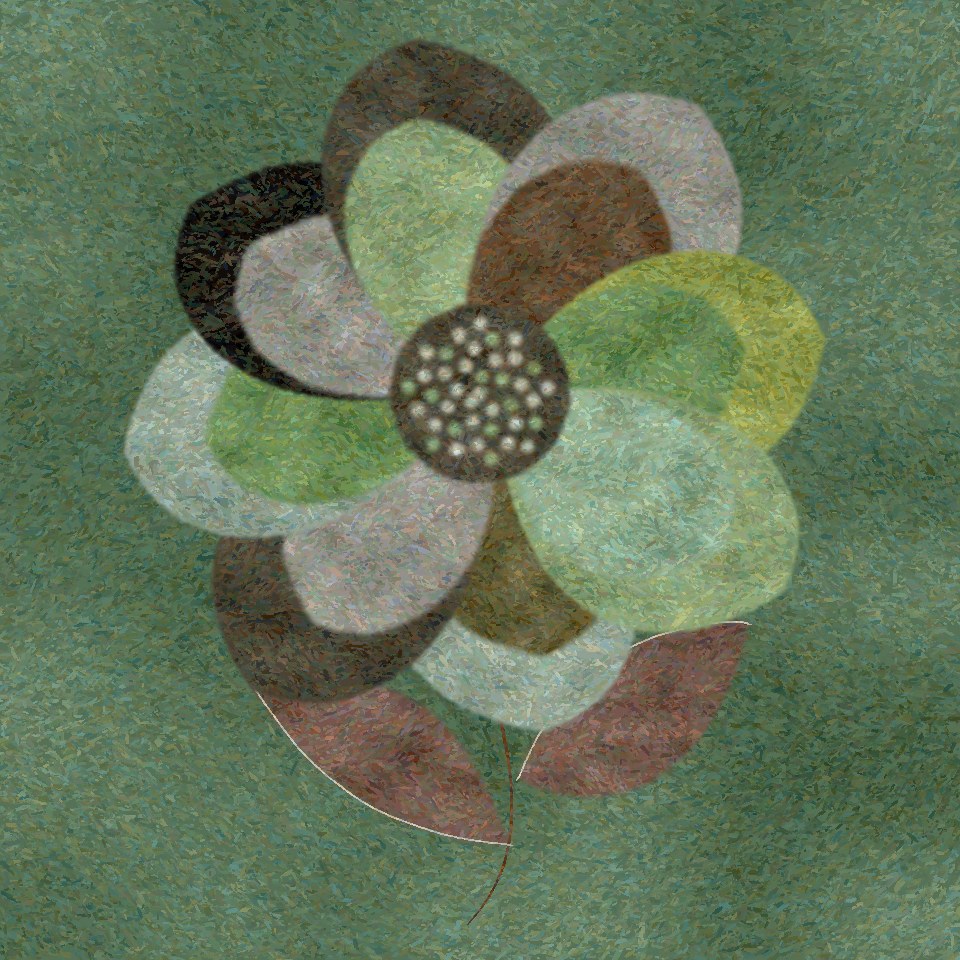} &
\includegraphics[width=.12\textwidth]{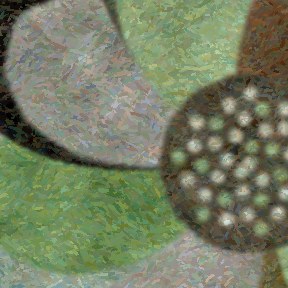} &
\includegraphics[width=.12\textwidth]{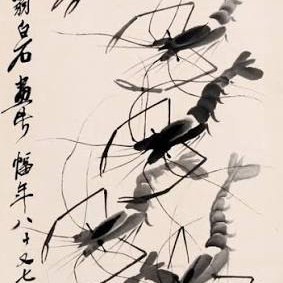} &
\includegraphics[width=.12\textwidth]{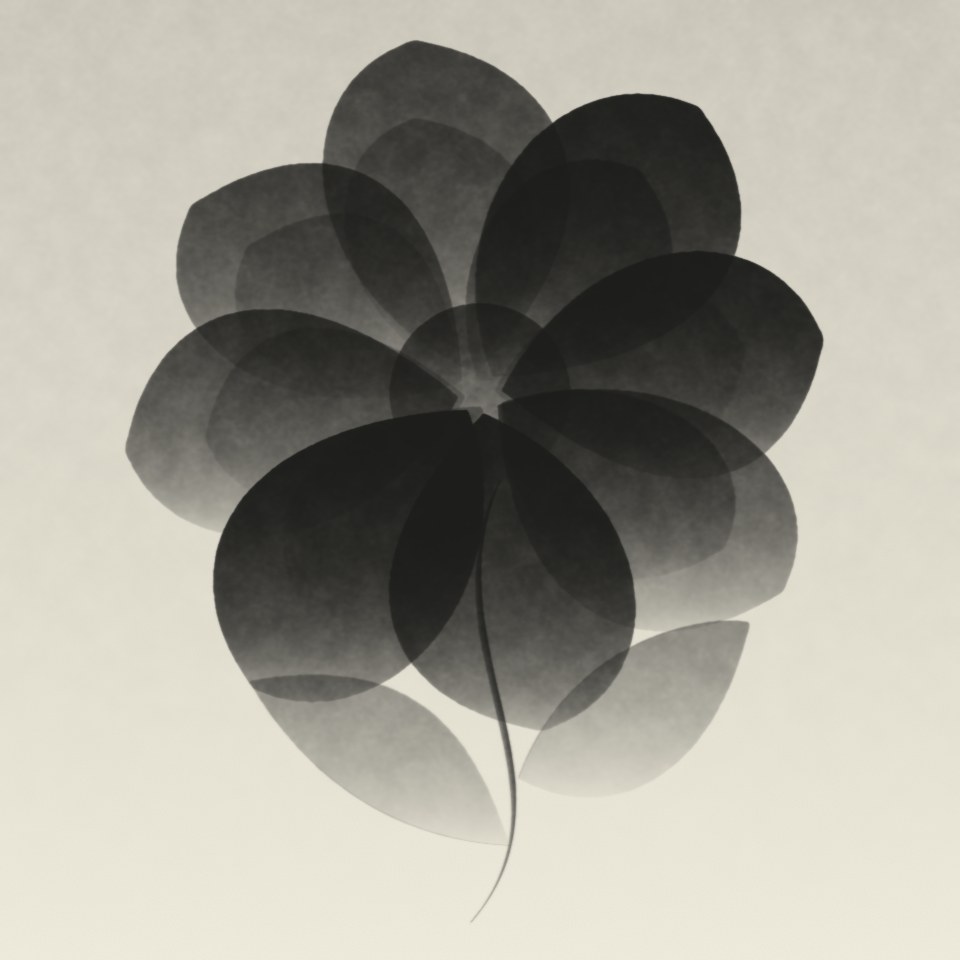} &
\includegraphics[width=.12\textwidth]{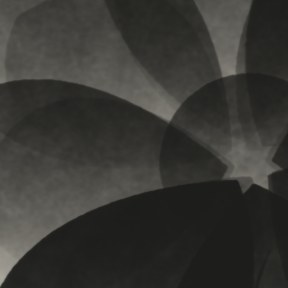}
\\[-1mm]
& \textsc{Monet} & & & \textsc{Qi Baishi} &
\\[3mm]

\includegraphics[width=.12\textwidth]{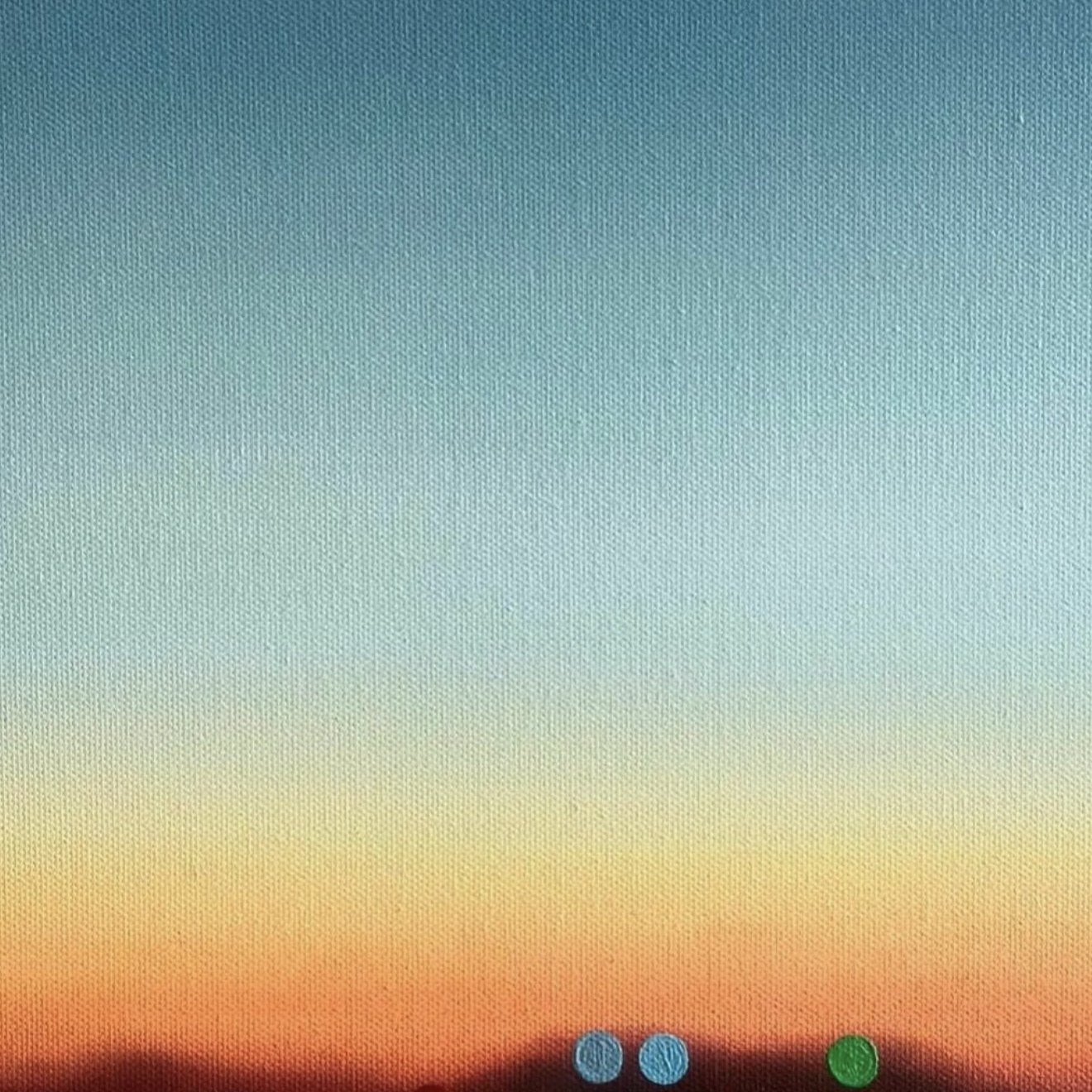} &
\includegraphics[width=.12\textwidth]{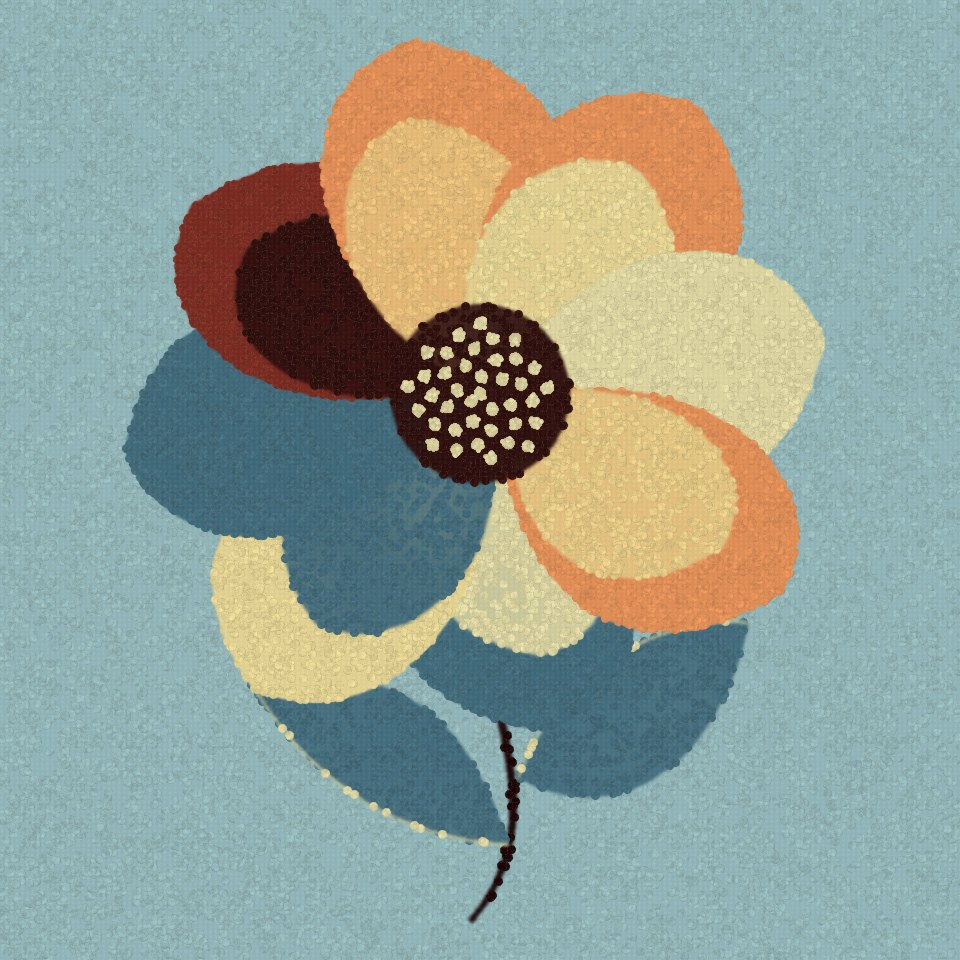} &
\includegraphics[width=.12\textwidth]{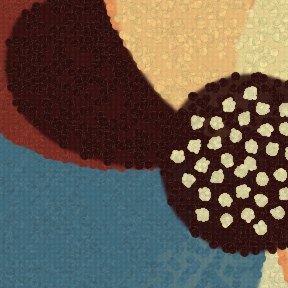} &
\includegraphics[width=.12\textwidth]{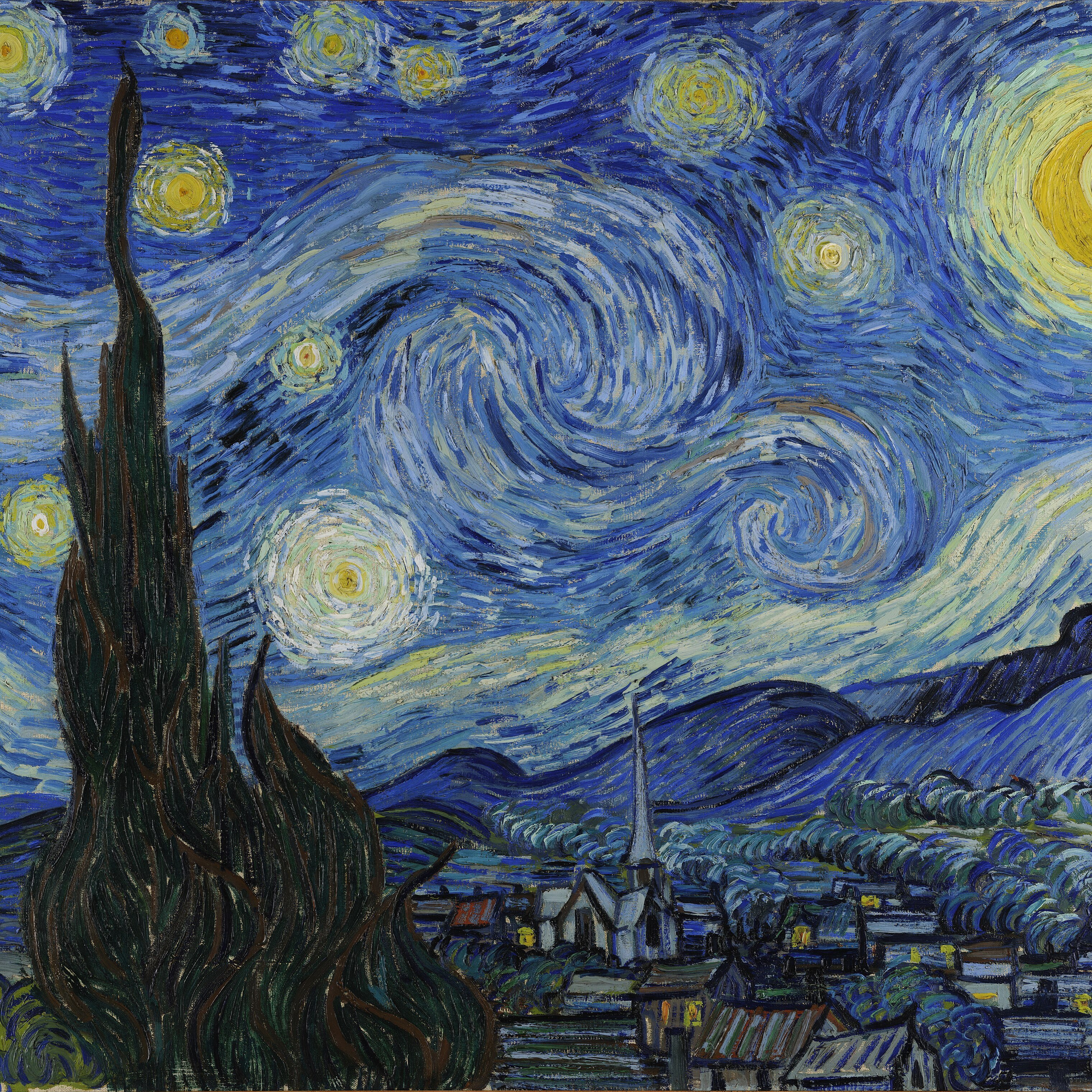} &
\includegraphics[width=.12\textwidth]{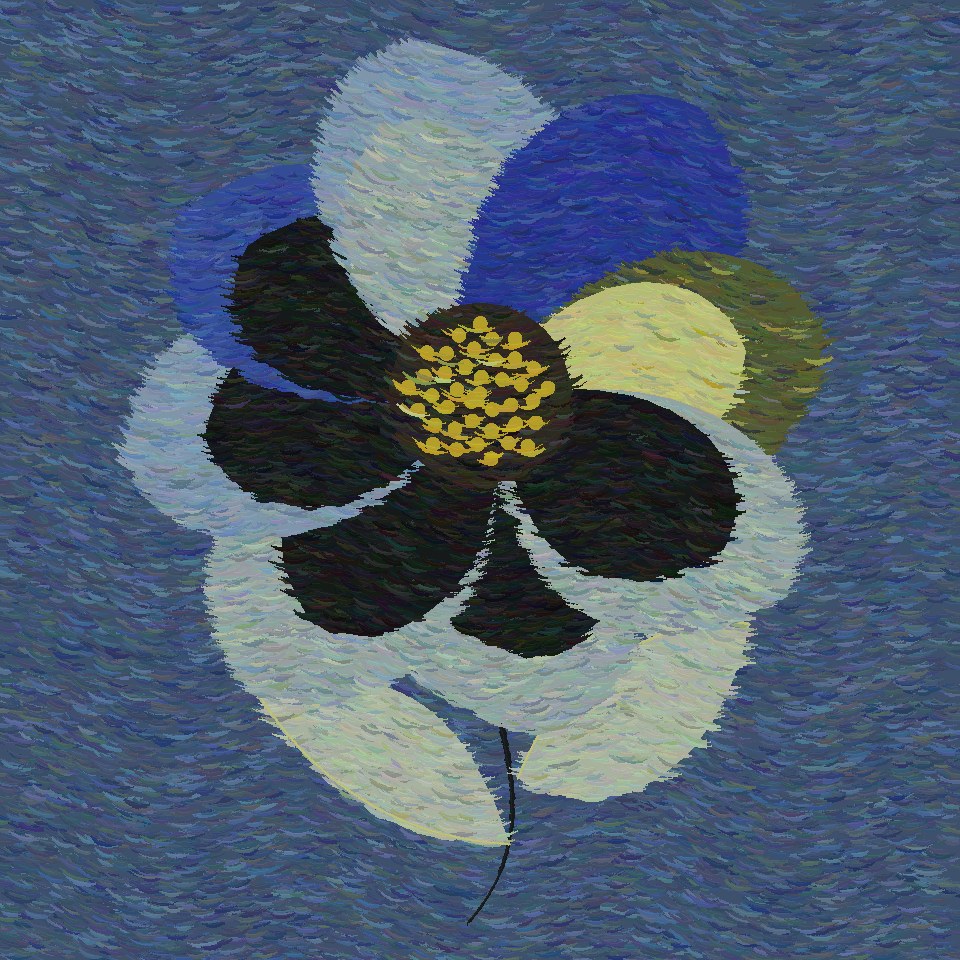} &
\includegraphics[width=.12\textwidth]{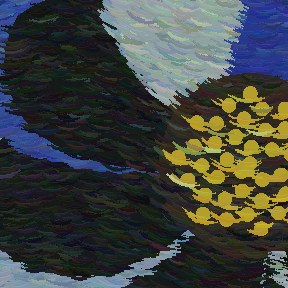}
\\[-1mm]
& \textsc{Riona} & & & \textsc{Van Gogh} &
\\[3mm]

\includegraphics[width=.12\textwidth]{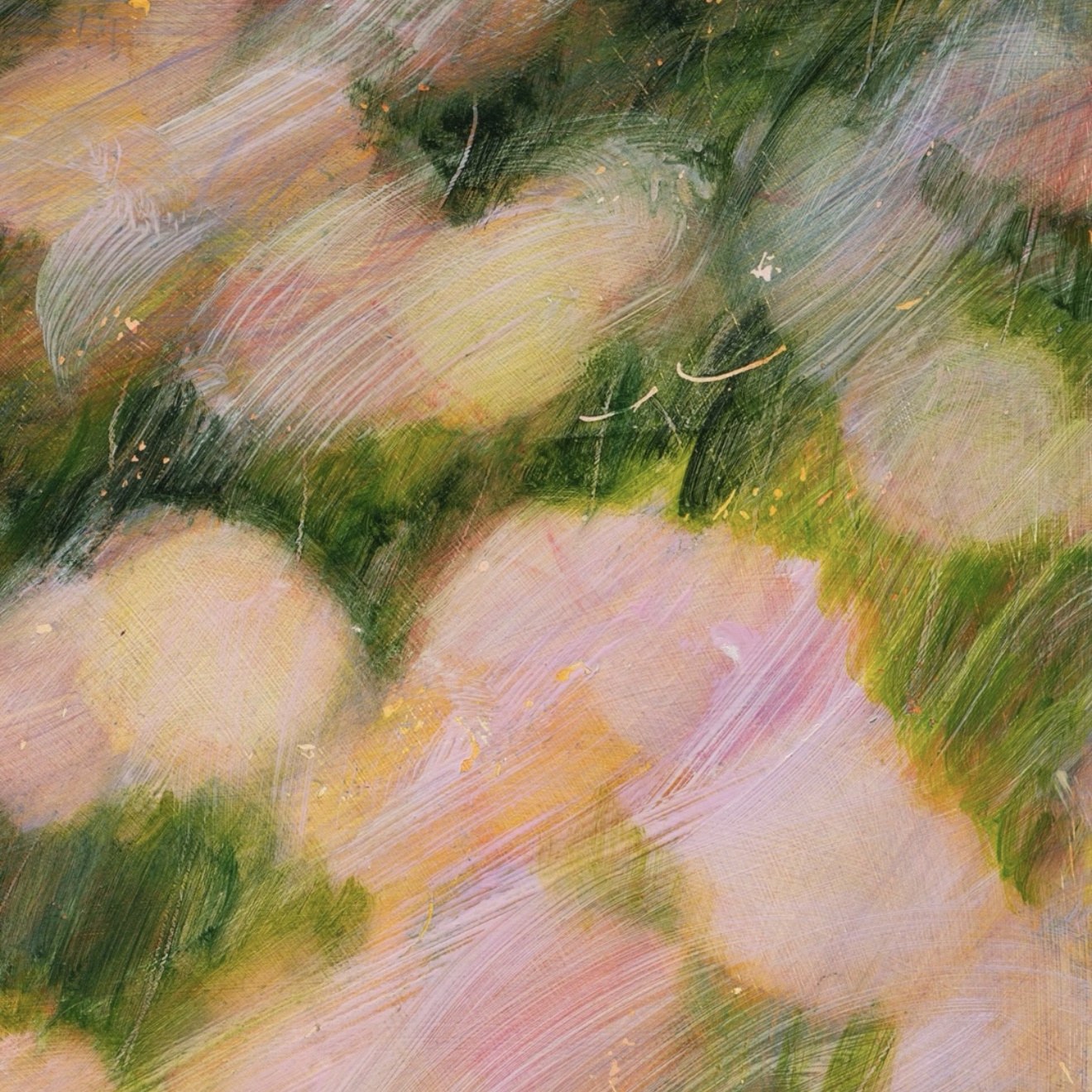} &
\includegraphics[width=.12\textwidth]{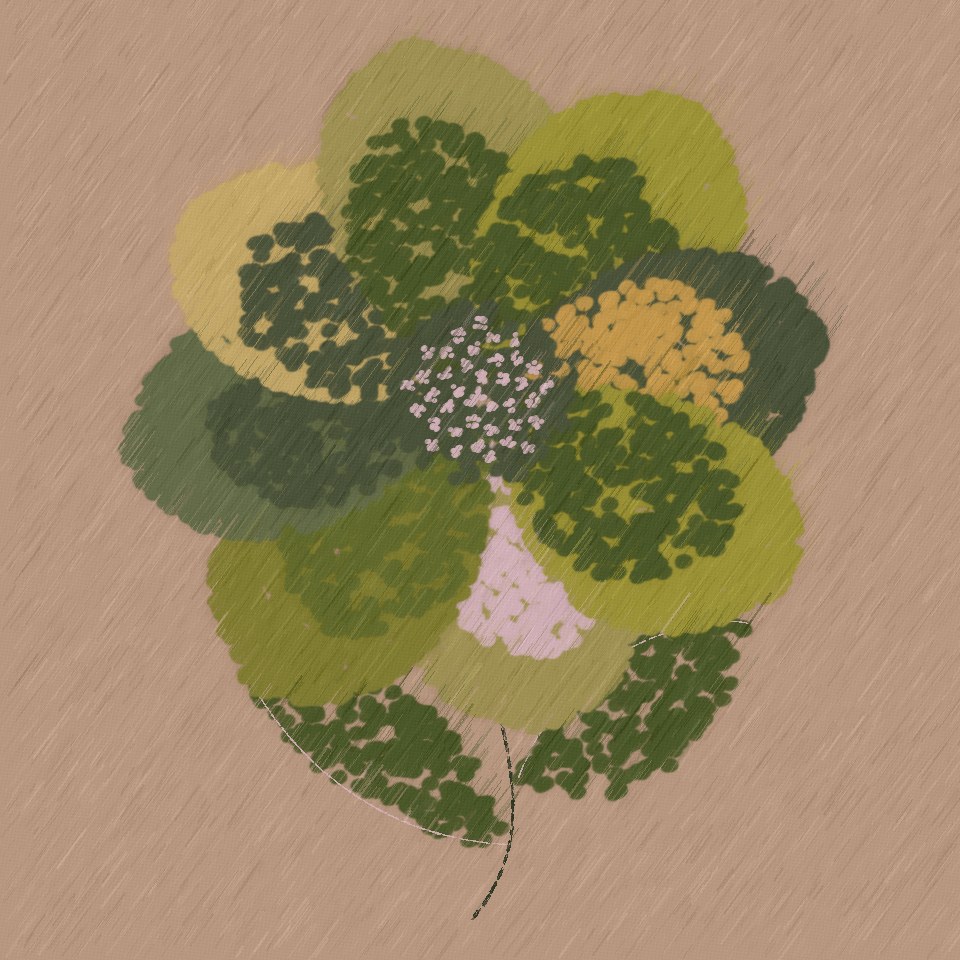} &
\includegraphics[width=.12\textwidth]{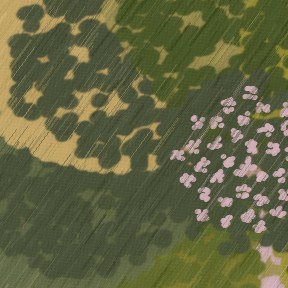} &
\includegraphics[width=.12\textwidth]{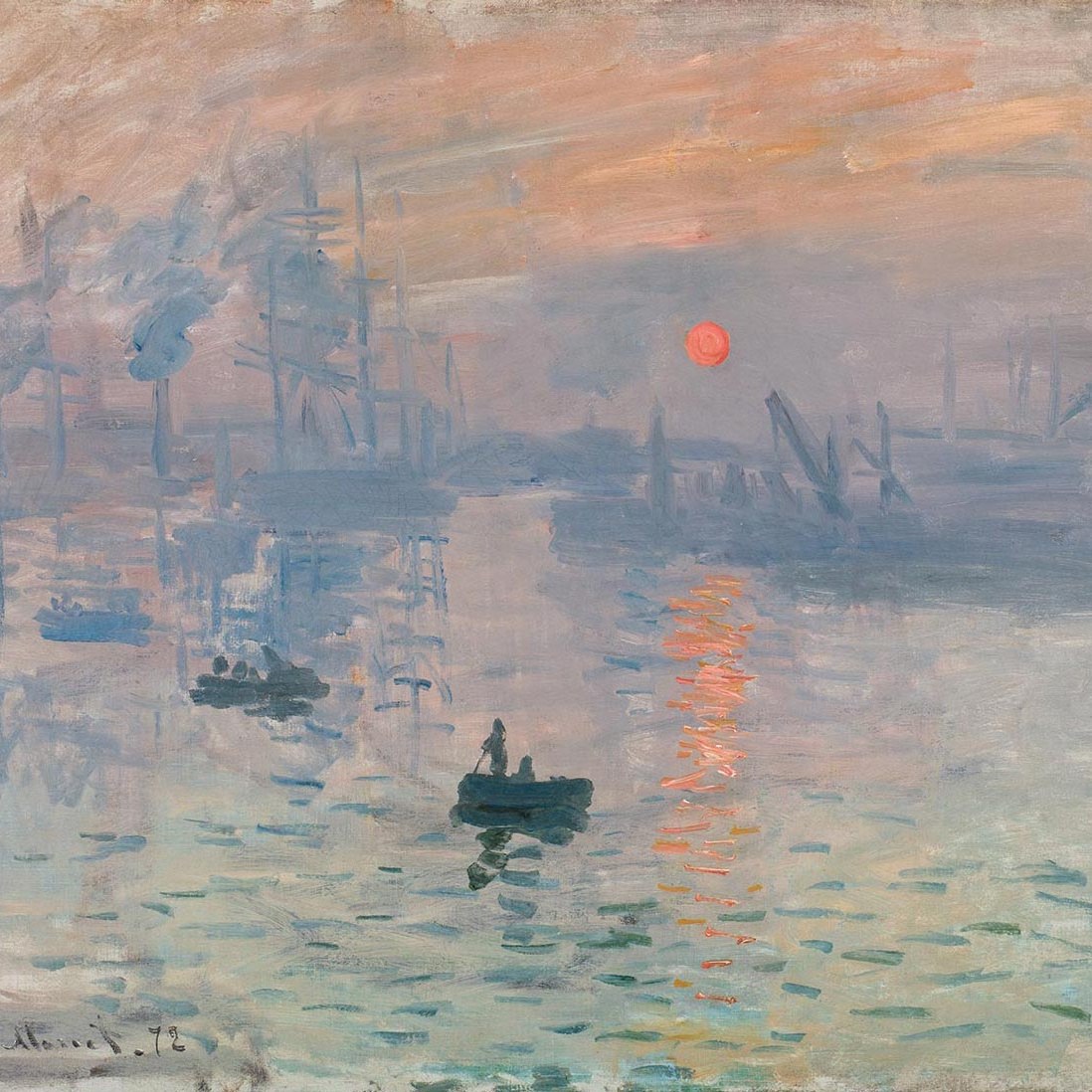} &
\includegraphics[width=.12\textwidth]{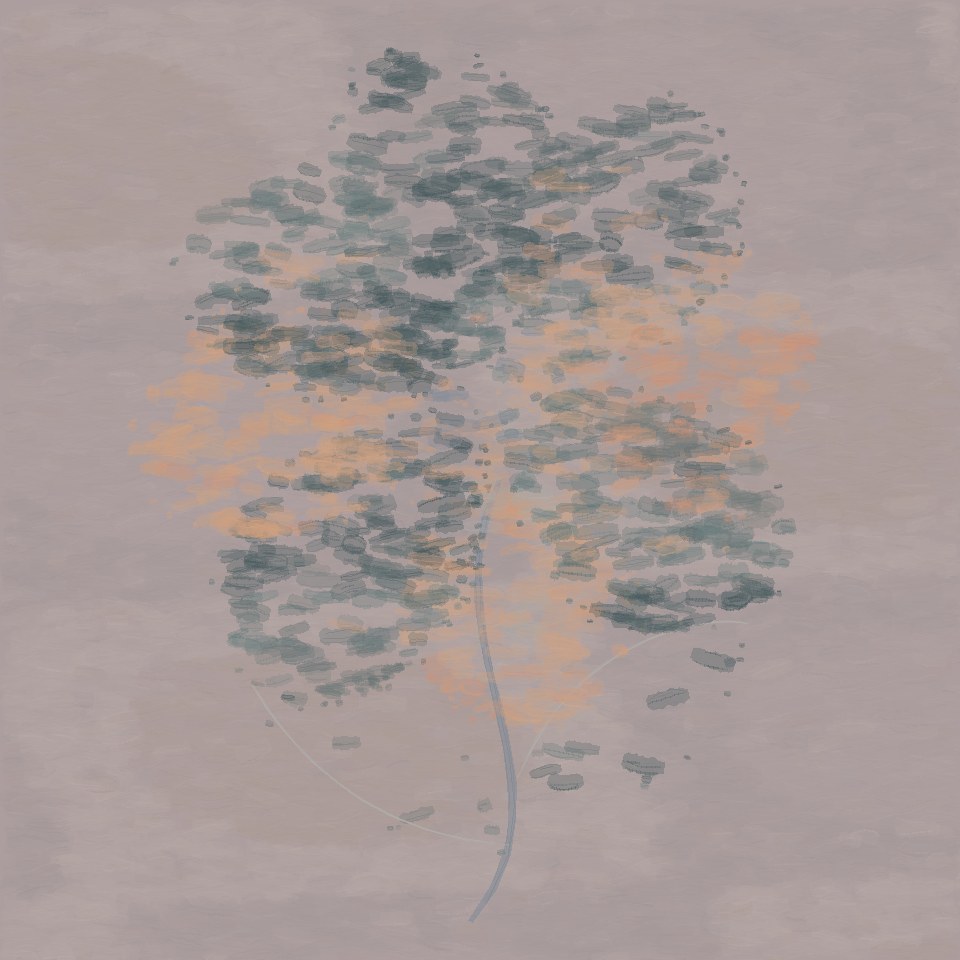} &
\includegraphics[width=.12\textwidth]{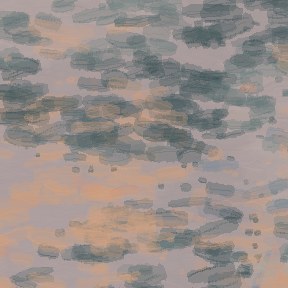}
\\[-1mm]
& \textsc{Smorczewski} & & & \textsc{Monet} &
\\[3mm]

\end{tabular}

\caption{
Fast style adaptation of a shared flower sketch. 
For each style, the reference artwork, the complete rendered flower, and an enlarged brush-detail crop are shown from left to right.
}

\label{fig:flower}
\end{figure*}

\begin{figure*}[t]
\centering
\setlength{\tabcolsep}{4pt}
\renewcommand{\arraystretch}{1.05}

\begin{tabular}{cccc}

\includegraphics[width=.12\textwidth]{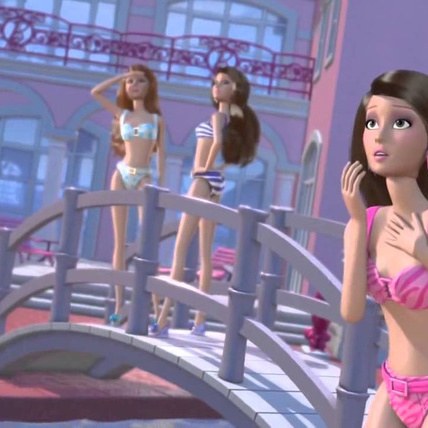} &
\includegraphics[width=.22\textwidth]{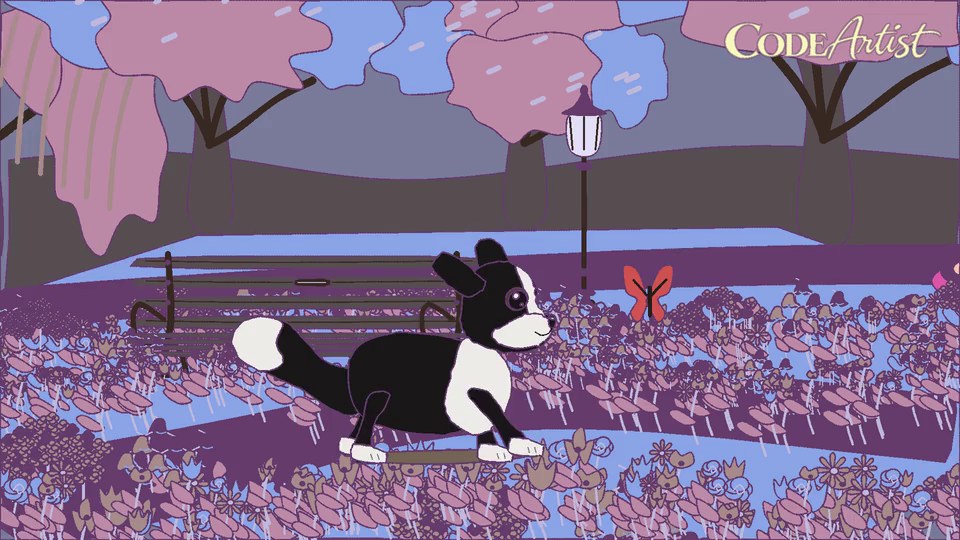} &
\includegraphics[width=.12\textwidth]{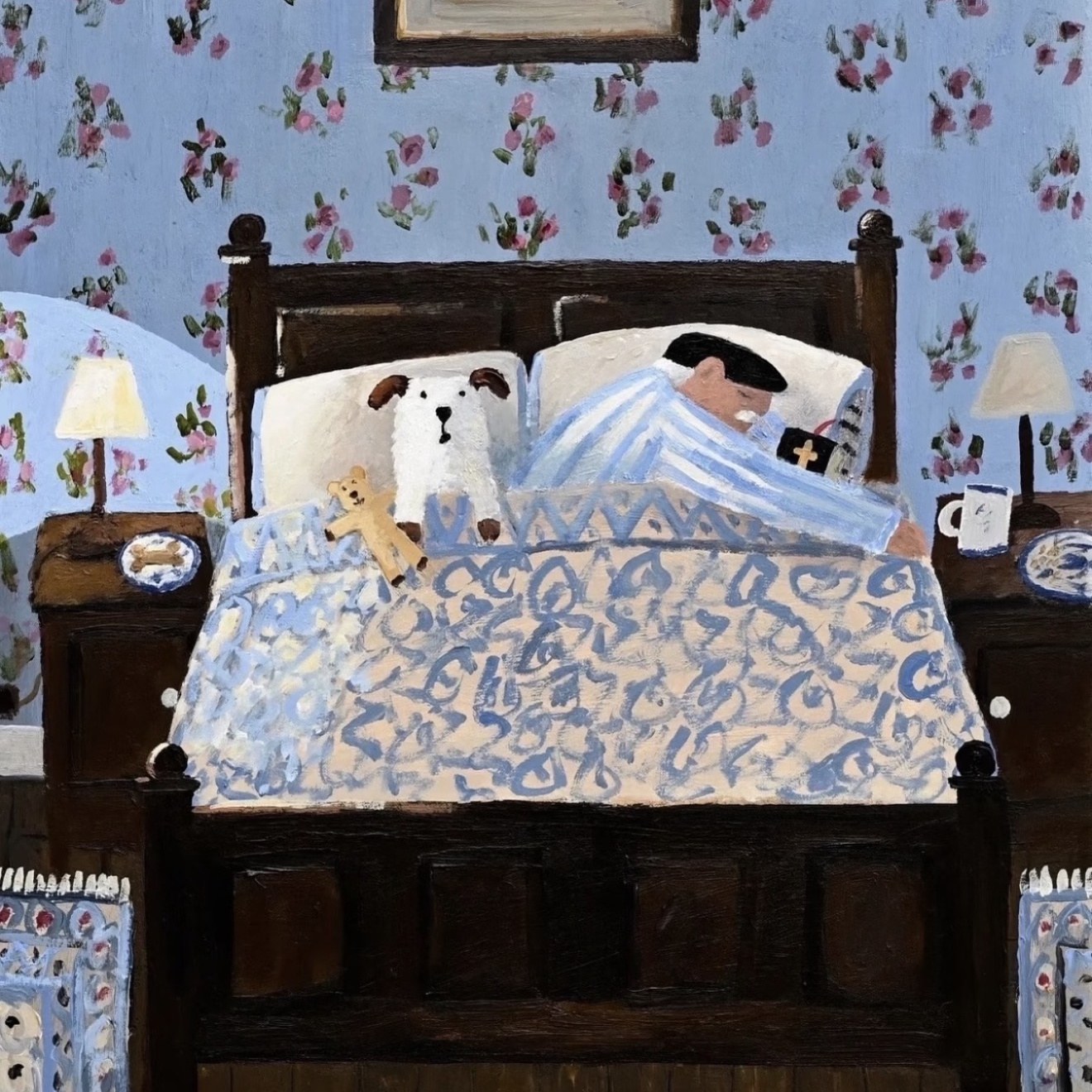} &
\includegraphics[width=.22\textwidth]{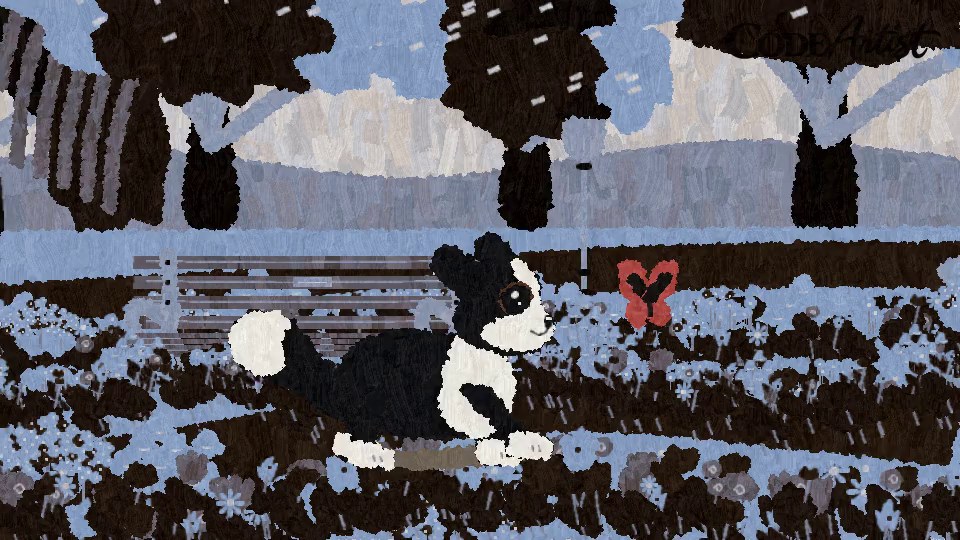}
\\[-1mm]
& \textsc{Barbie} & & \textsc{Gary}
\\[3mm]

\includegraphics[width=.12\textwidth]{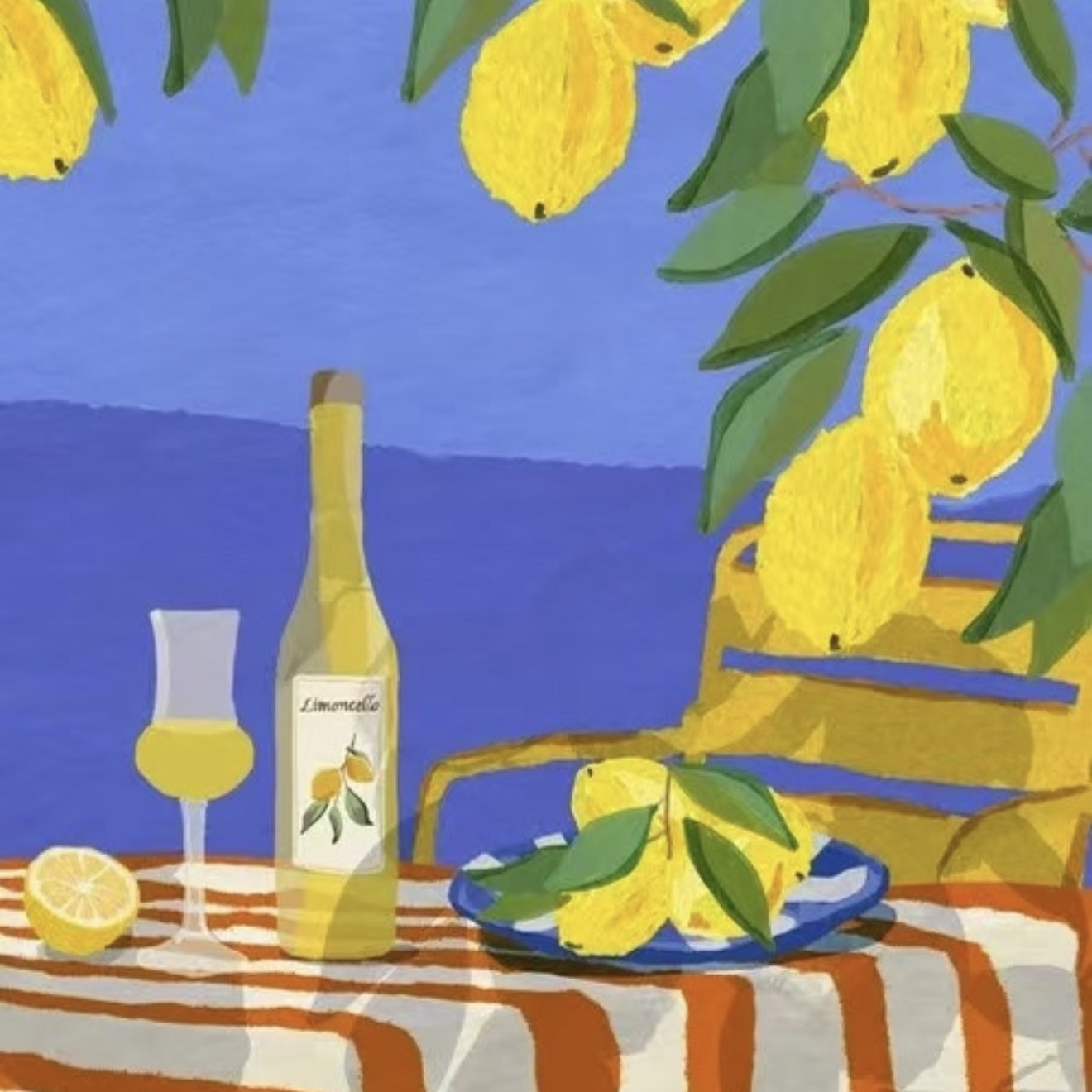} &
\includegraphics[width=.22\textwidth]{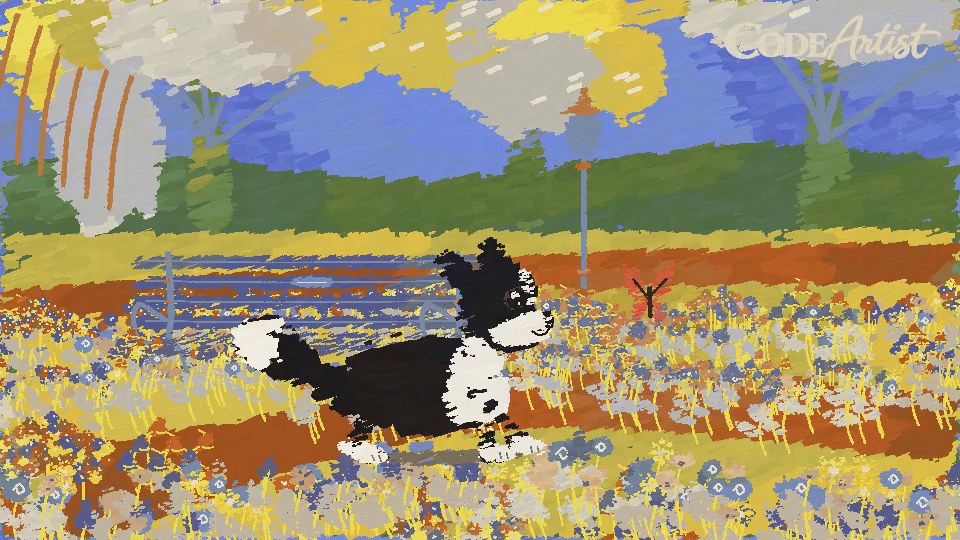} &
\includegraphics[width=.12\textwidth]{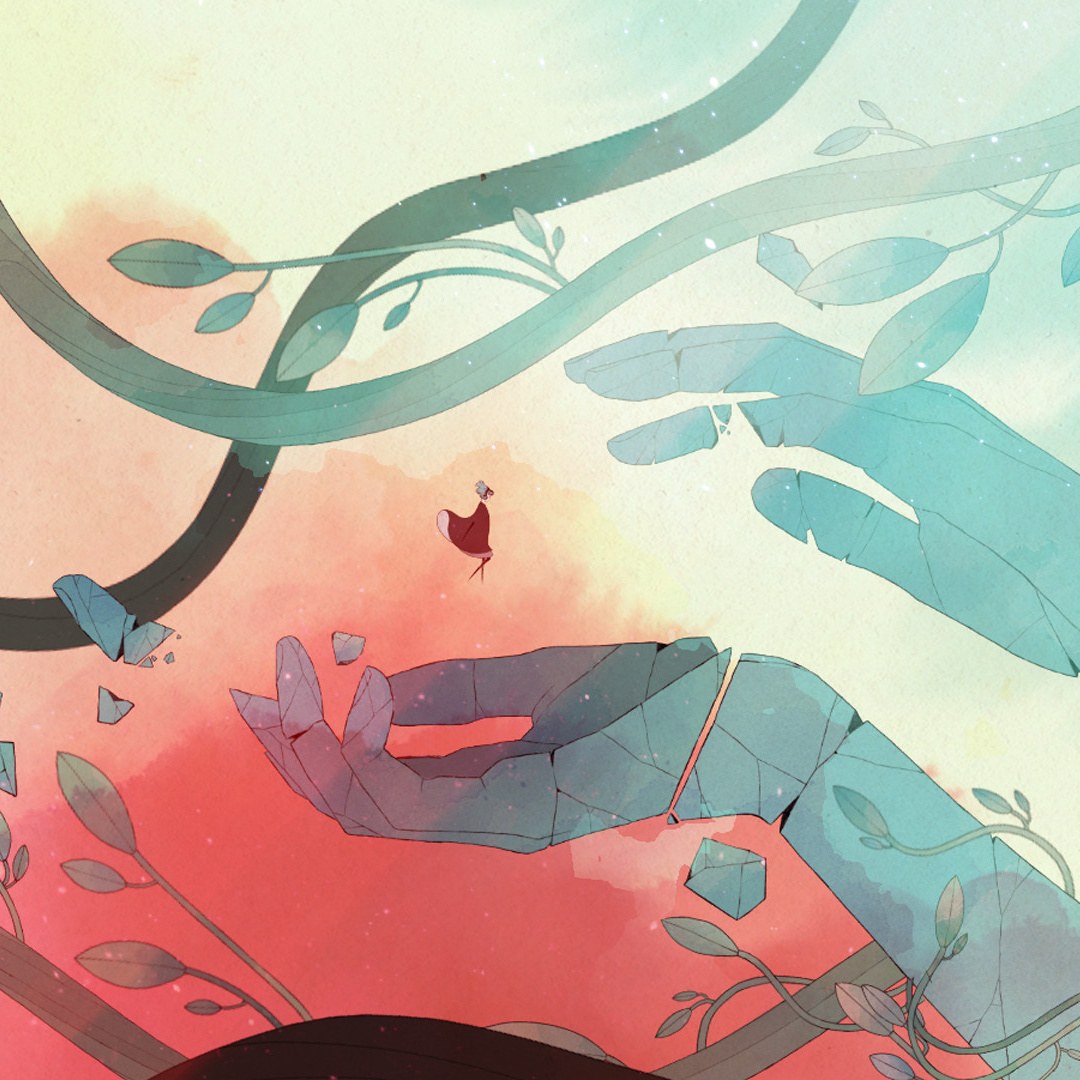} &
\includegraphics[width=.22\textwidth]{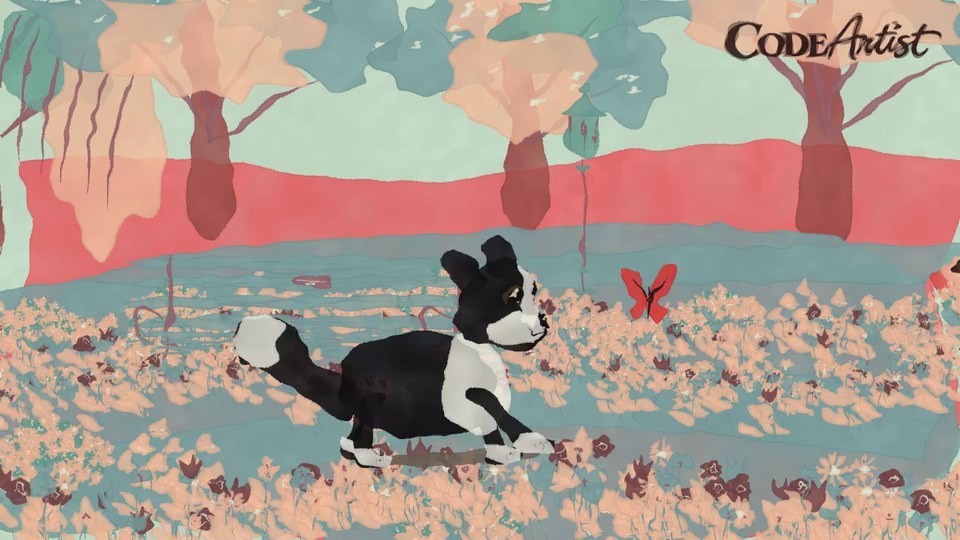}
\\[-1mm]
& \textsc{Catherine} & & \textsc{Gris}
\\[3mm]

\includegraphics[width=.12\textwidth]{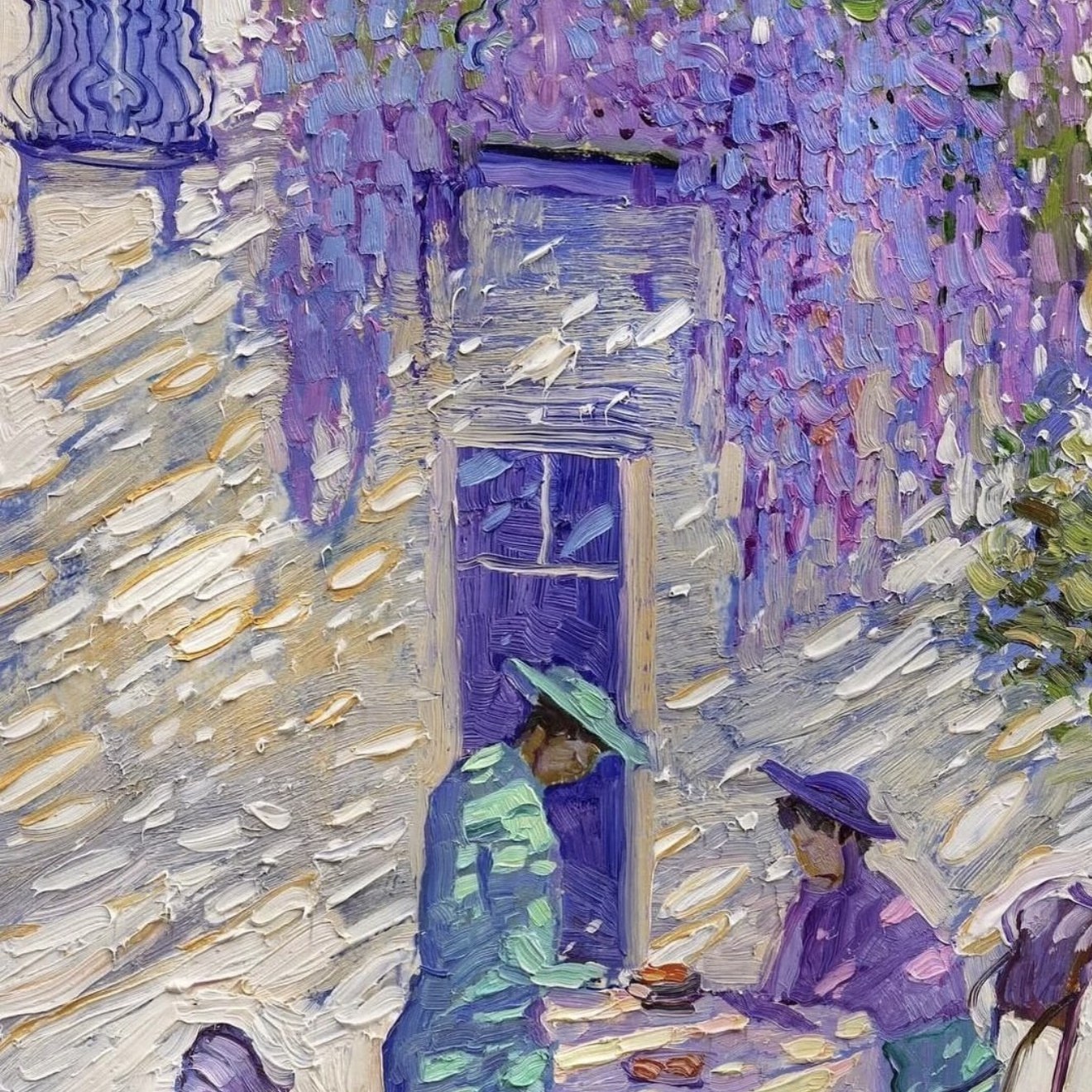} &
\includegraphics[width=.22\textwidth]{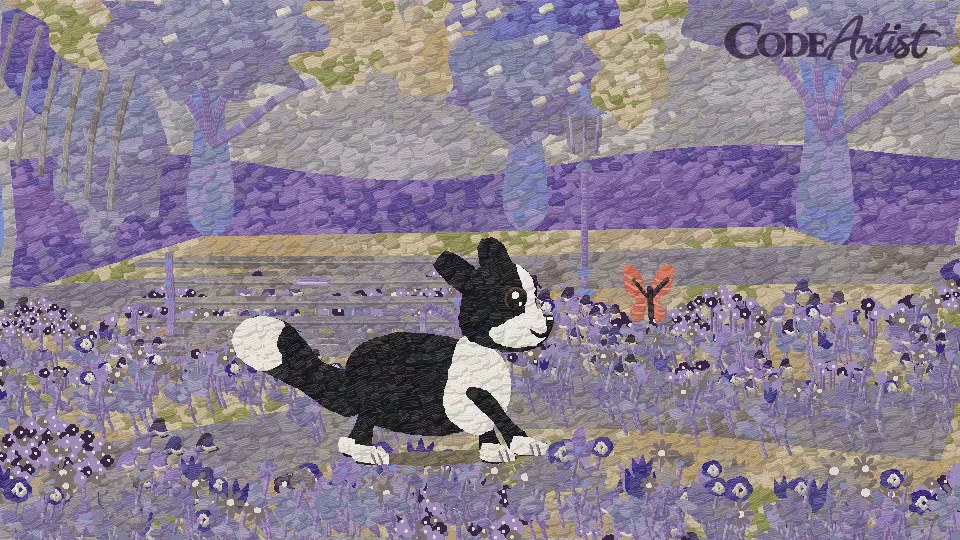} &
\includegraphics[width=.12\textwidth]{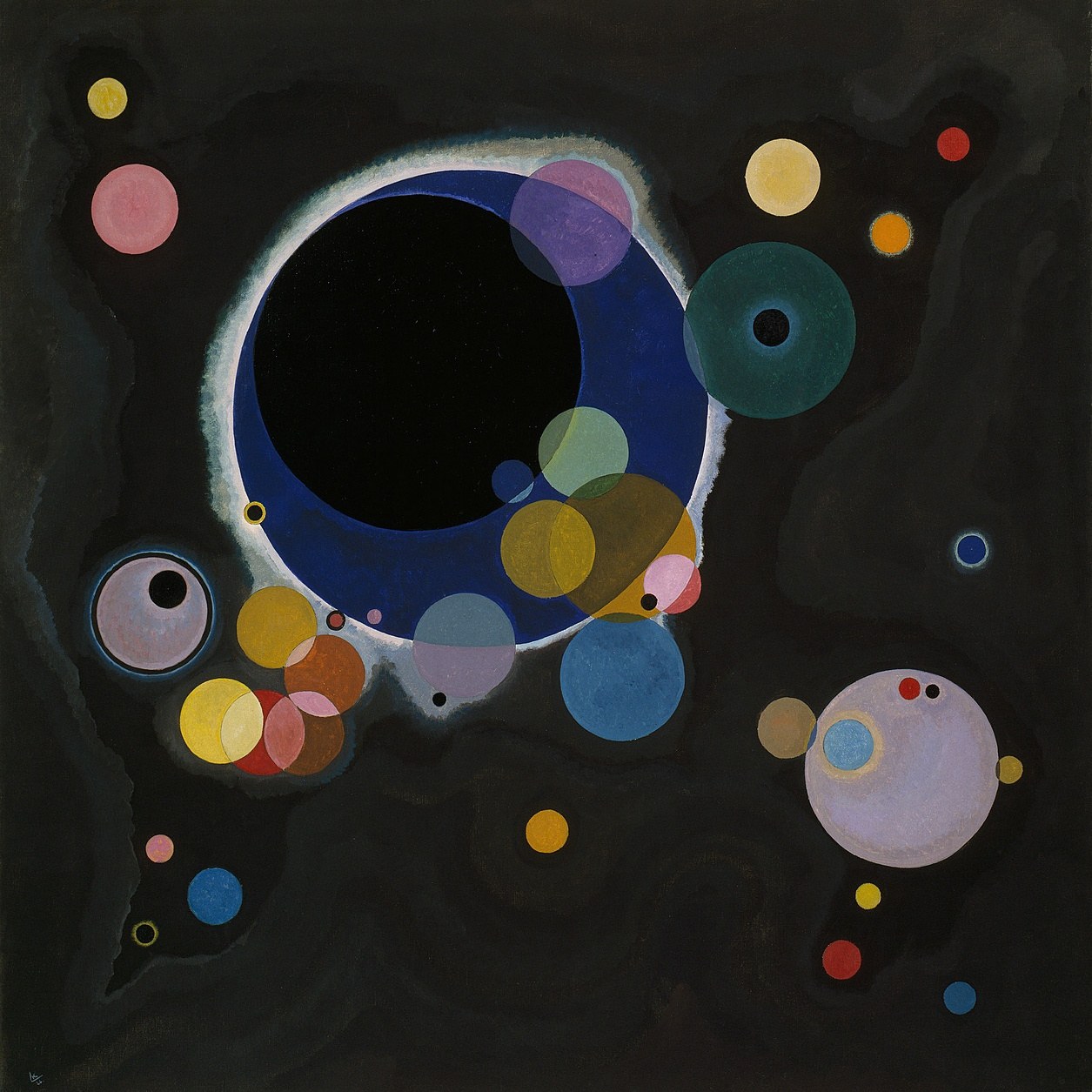} &
\includegraphics[width=.22\textwidth]{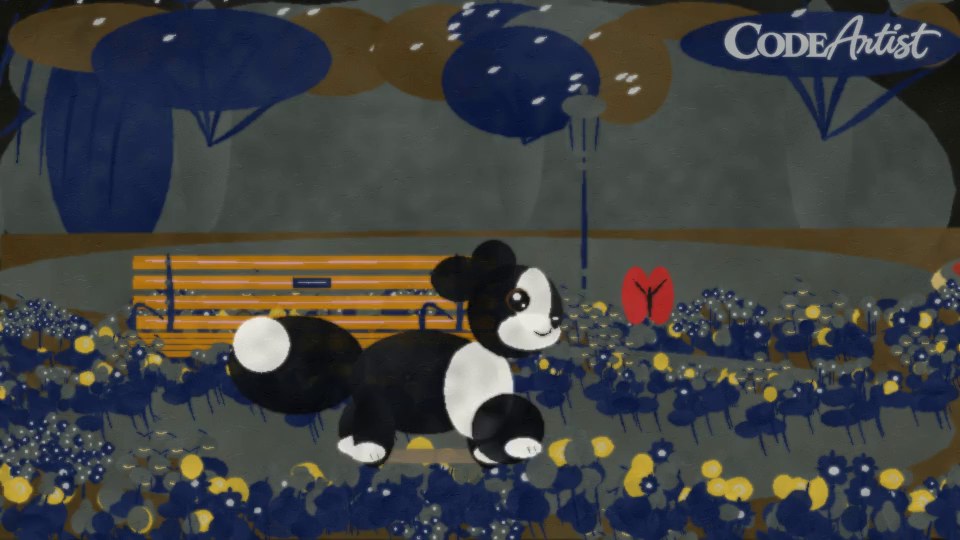}
\\[-1mm]
& \textsc{Jacaranda} & & \textsc{Kandinsky}
\\[3mm]

\includegraphics[width=.12\textwidth]{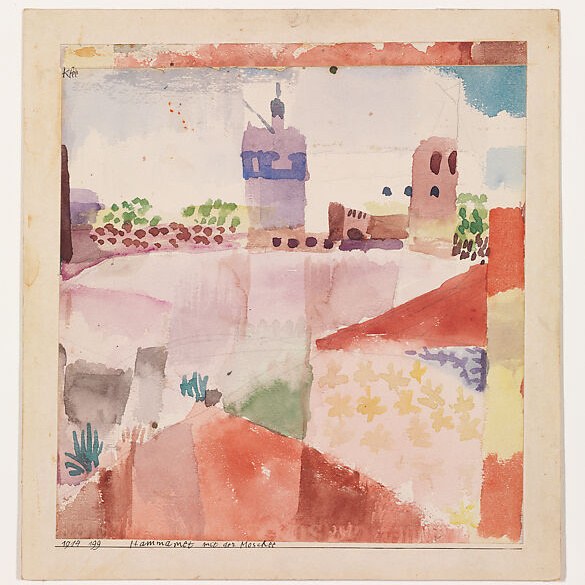} &
\includegraphics[width=.22\textwidth]{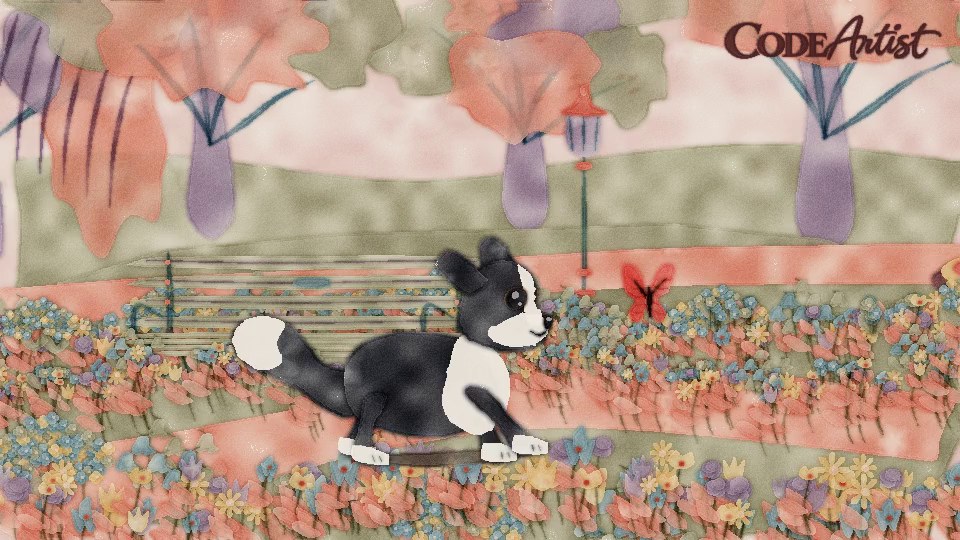} &
\includegraphics[width=.12\textwidth]{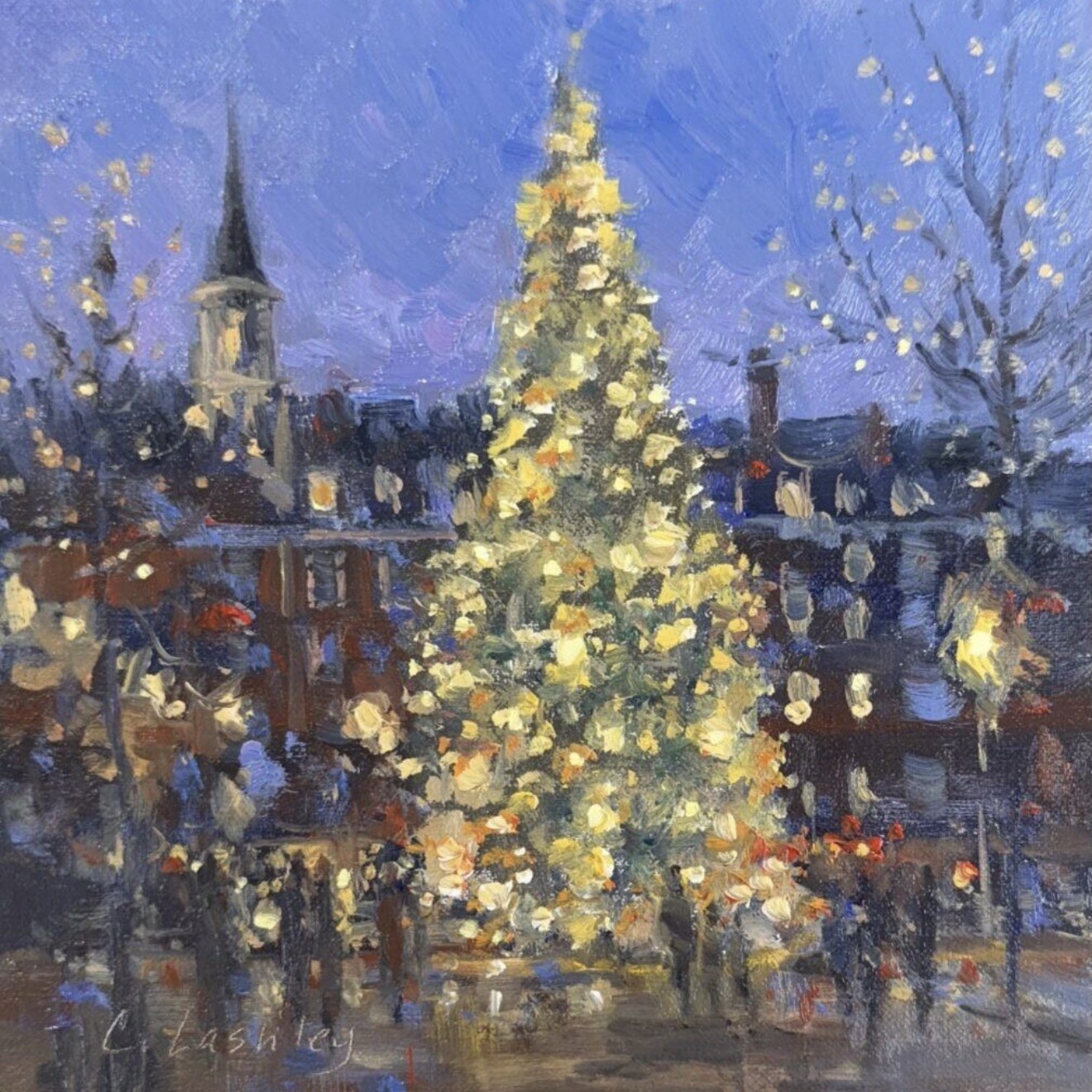} &
\includegraphics[width=.22\textwidth]{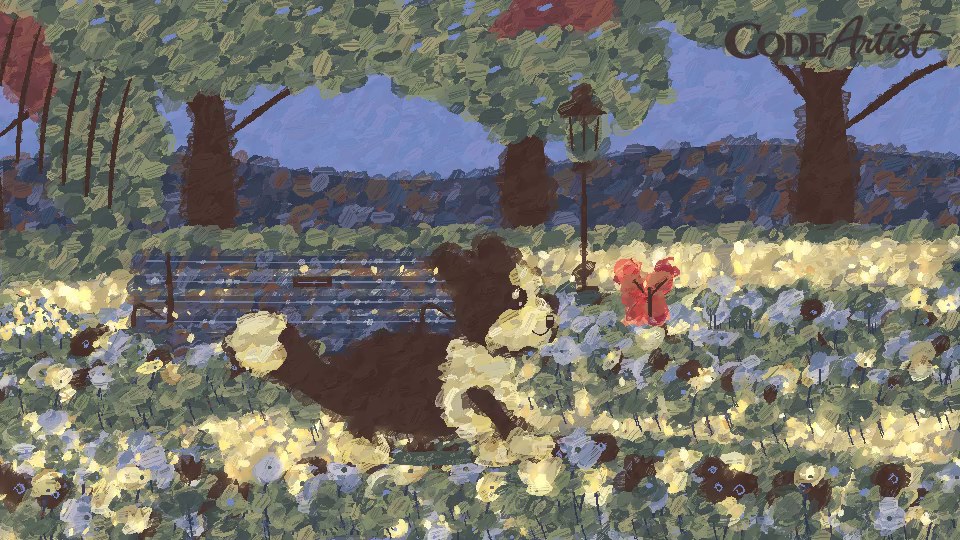}
\\[-1mm]
& \textsc{Klee} & & \textsc{Lashley}
\\[3mm]

\includegraphics[width=.12\textwidth]{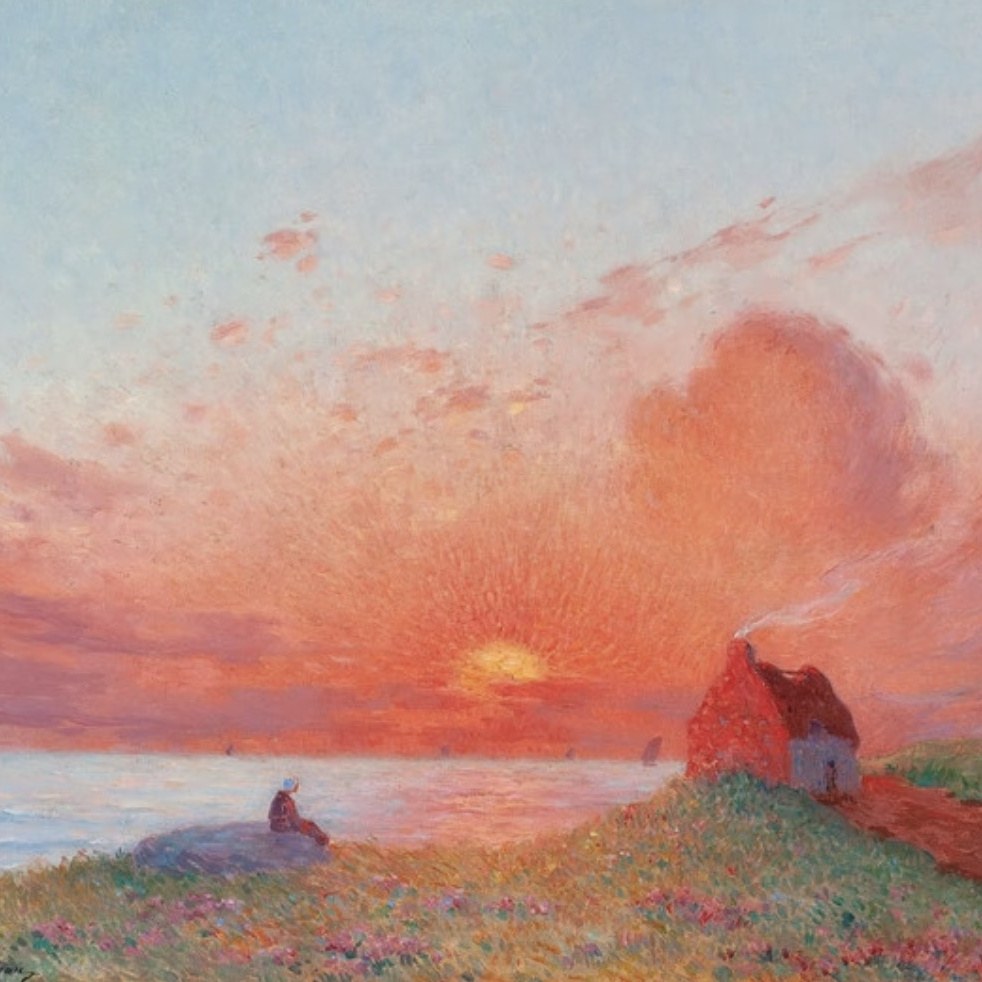} &
\includegraphics[width=.22\textwidth]{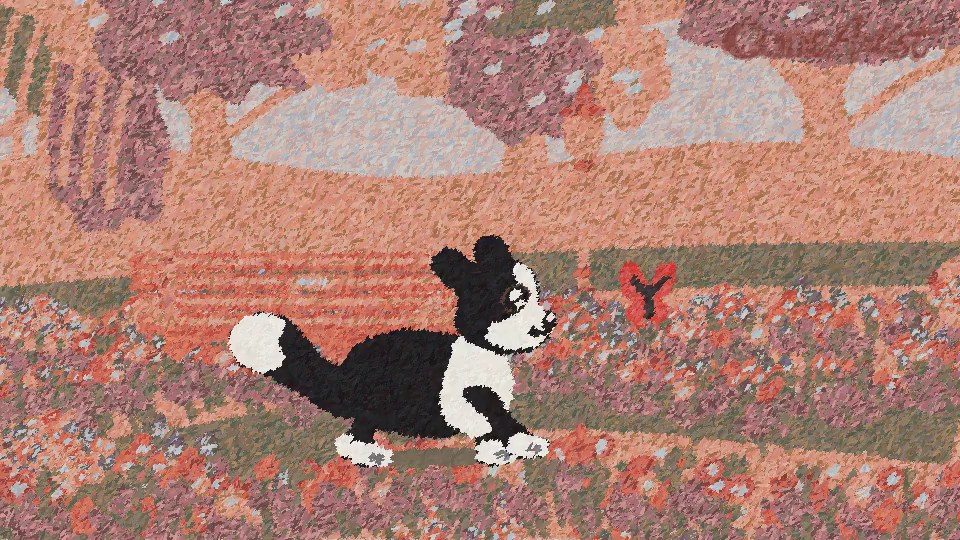} &
\includegraphics[width=.12\textwidth]{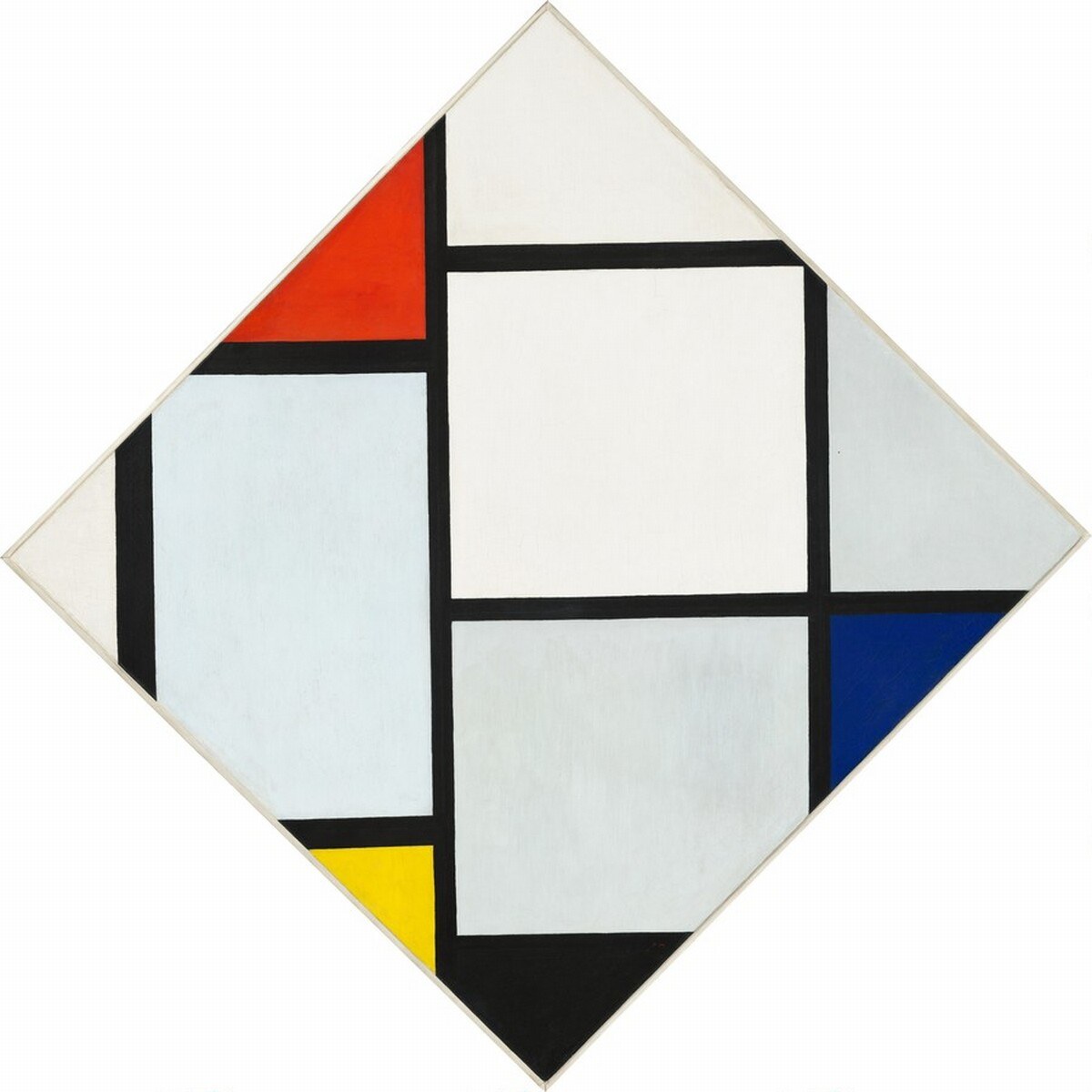} &
\includegraphics[width=.22\textwidth]{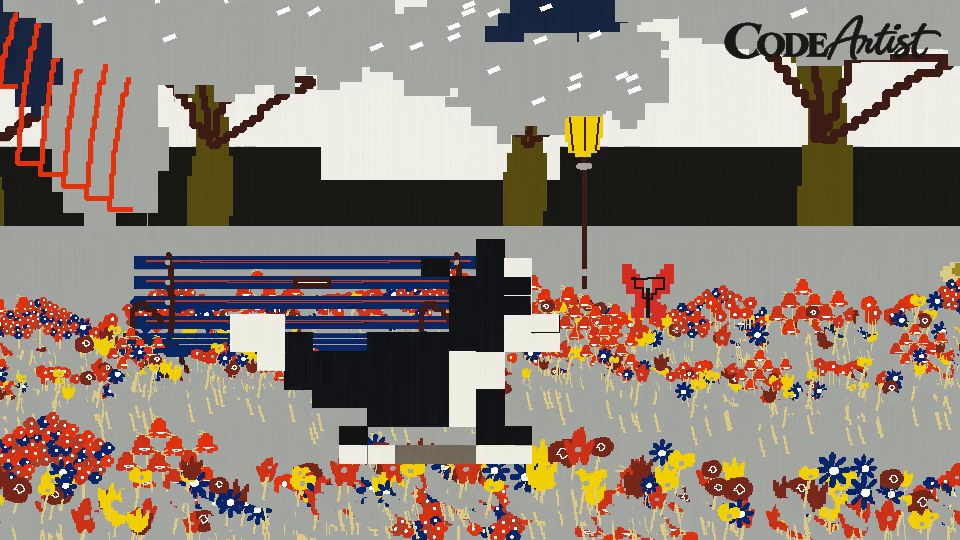}
\\[-1mm]
& \textsc{Lorn} & & \textsc{Mondrian}
\\[3mm]

\includegraphics[width=.12\textwidth]{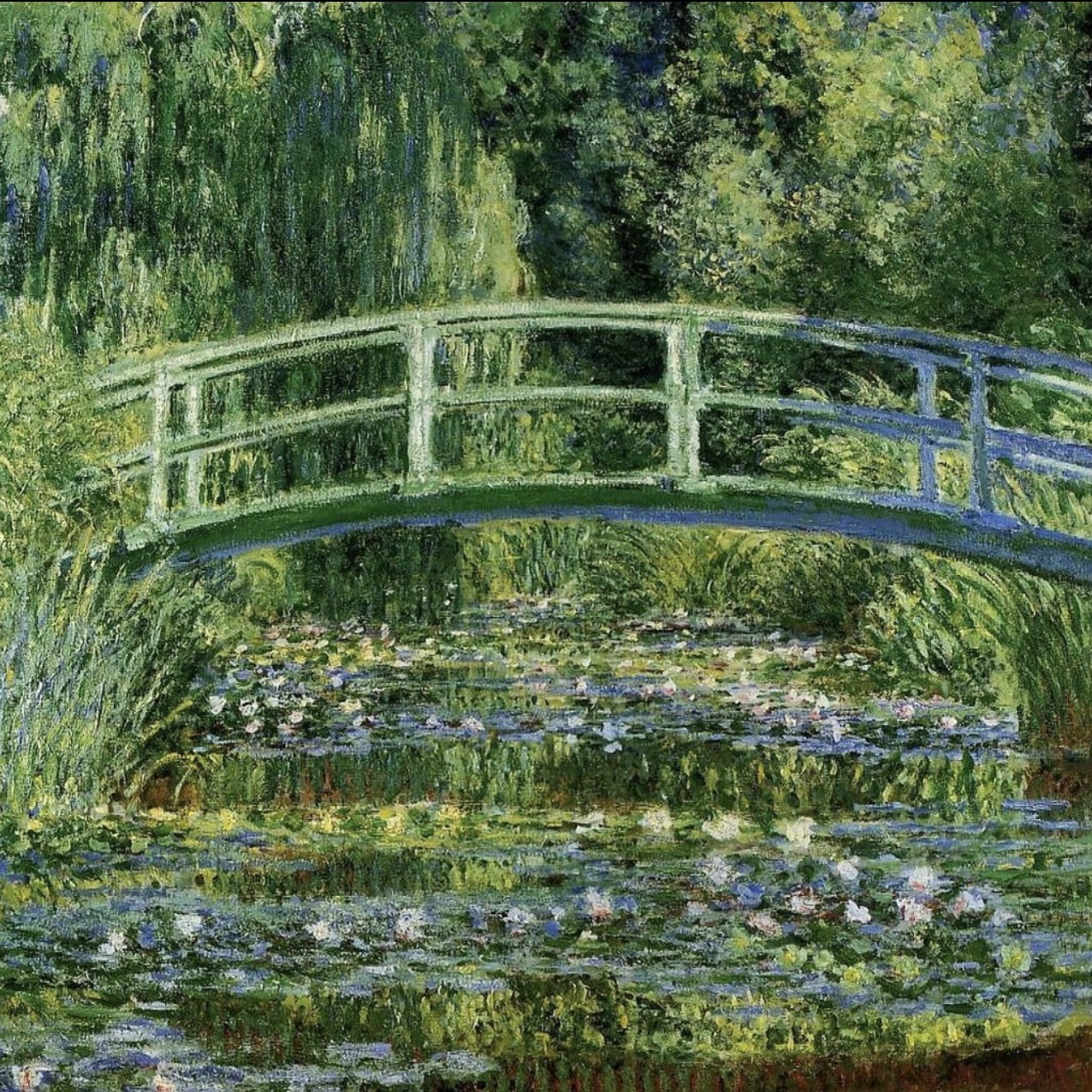} &
\includegraphics[width=.22\textwidth]{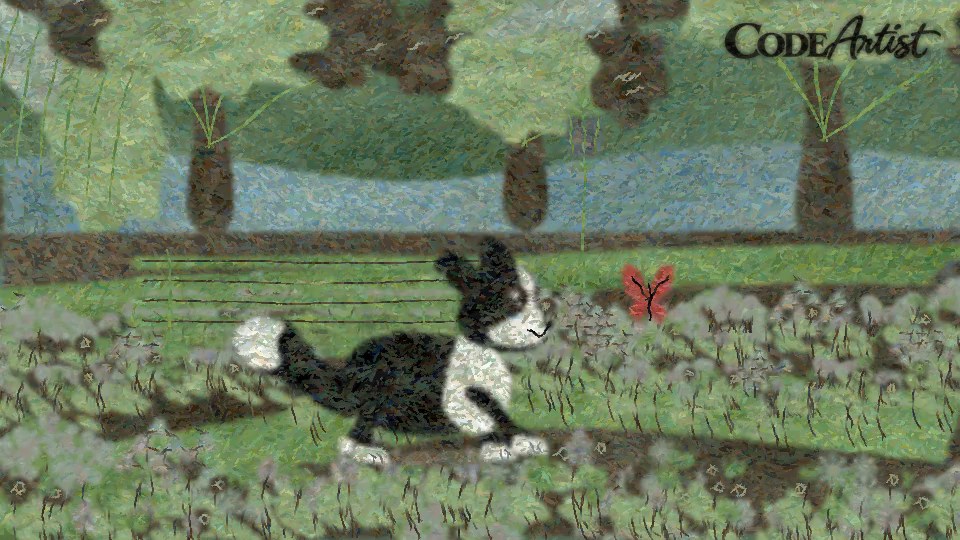} &
\includegraphics[width=.12\textwidth]{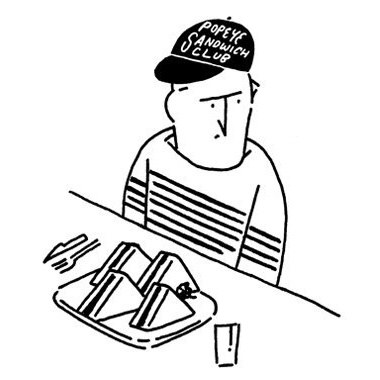} &
\includegraphics[width=.22\textwidth]{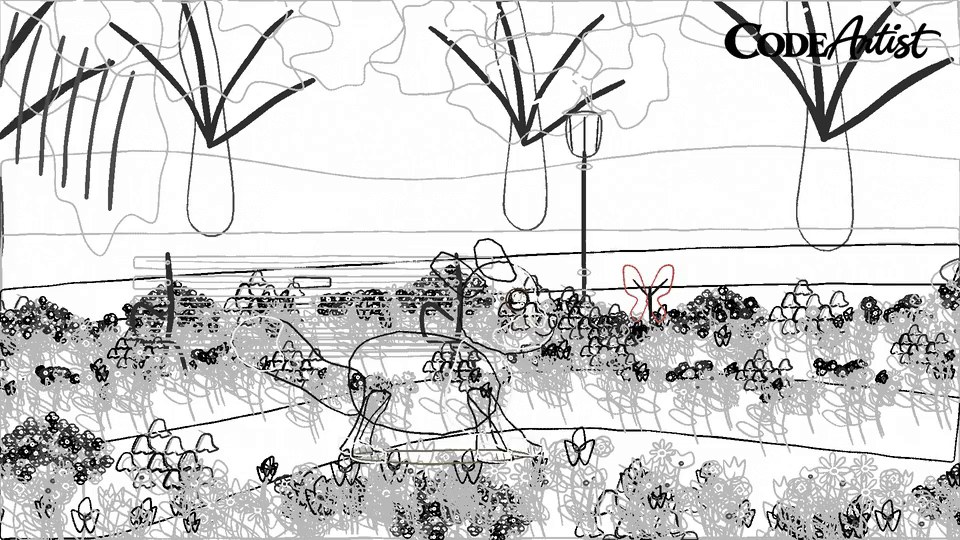}
\\[-1mm]
& \textsc{Monet} & & \textsc{Nagaba}
\\[3mm]

\includegraphics[width=.12\textwidth]{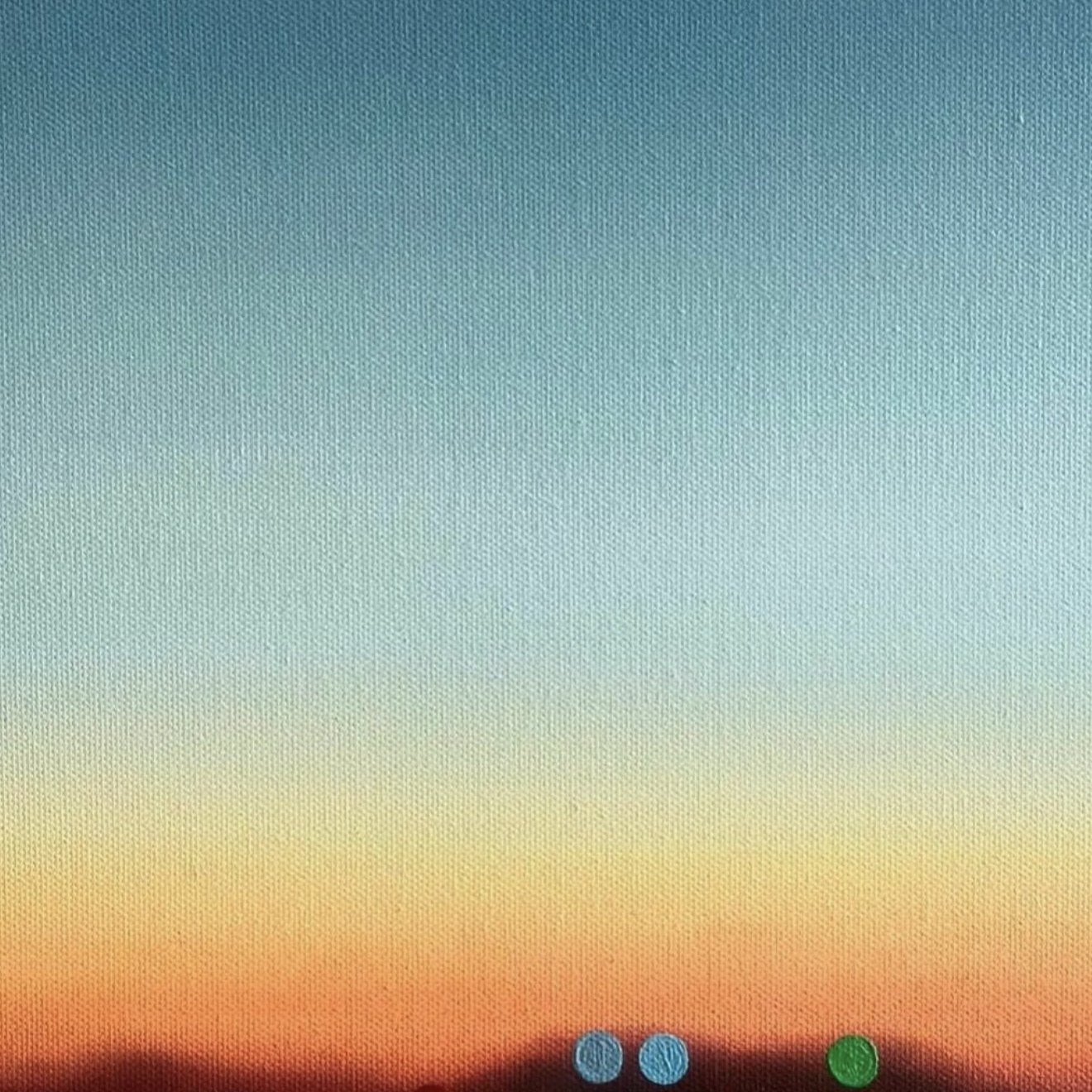} &
\includegraphics[width=.22\textwidth]{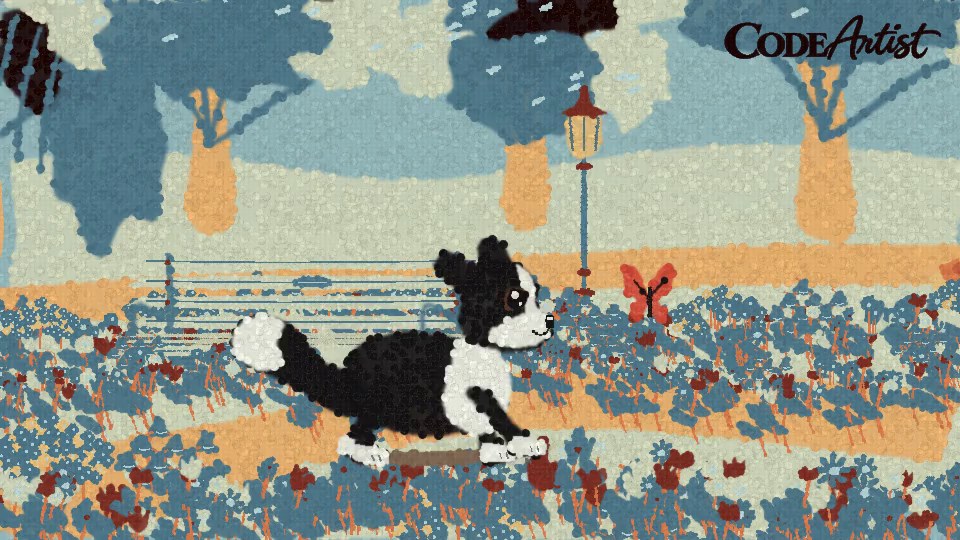} &
\includegraphics[width=.12\textwidth]{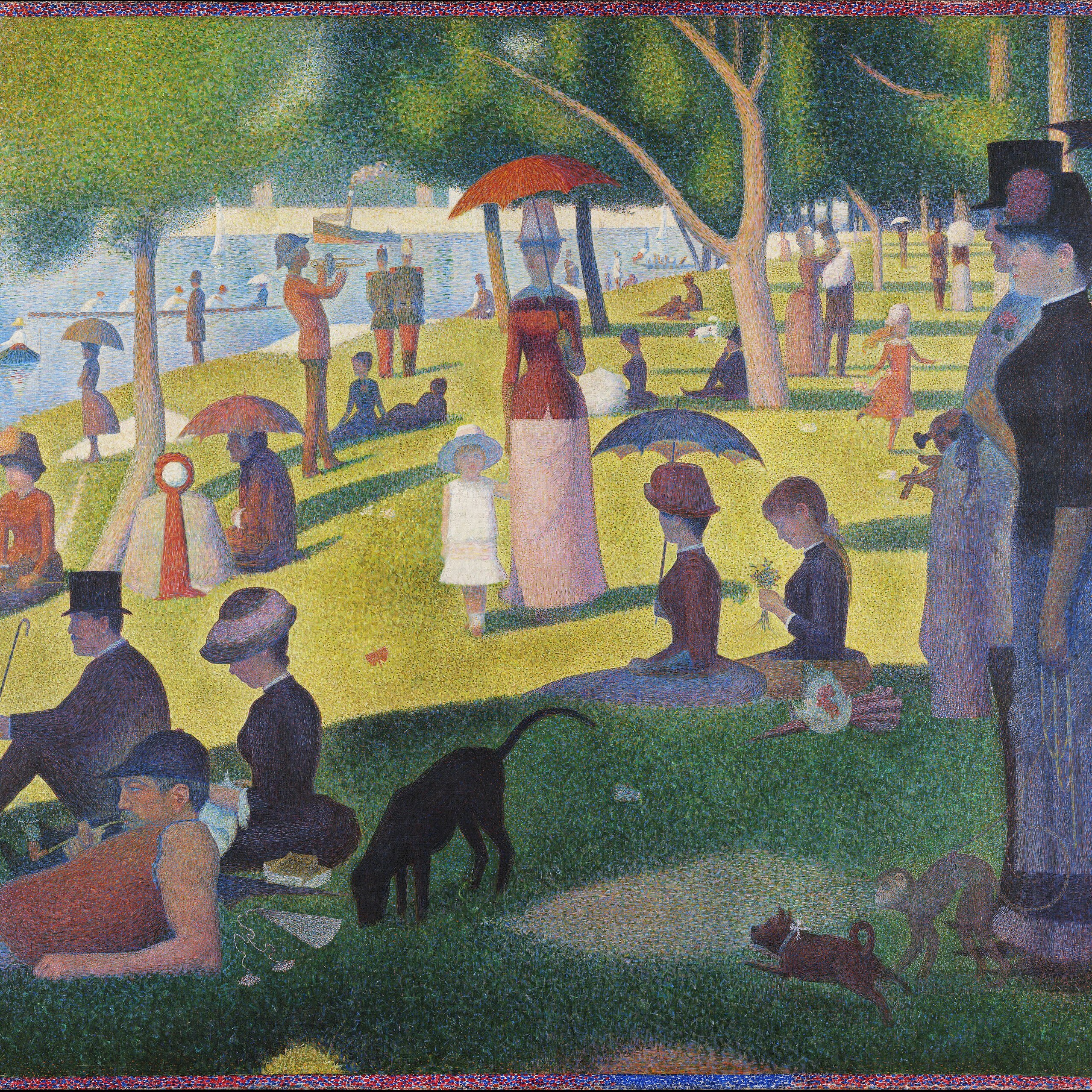} &
\includegraphics[width=.22\textwidth]{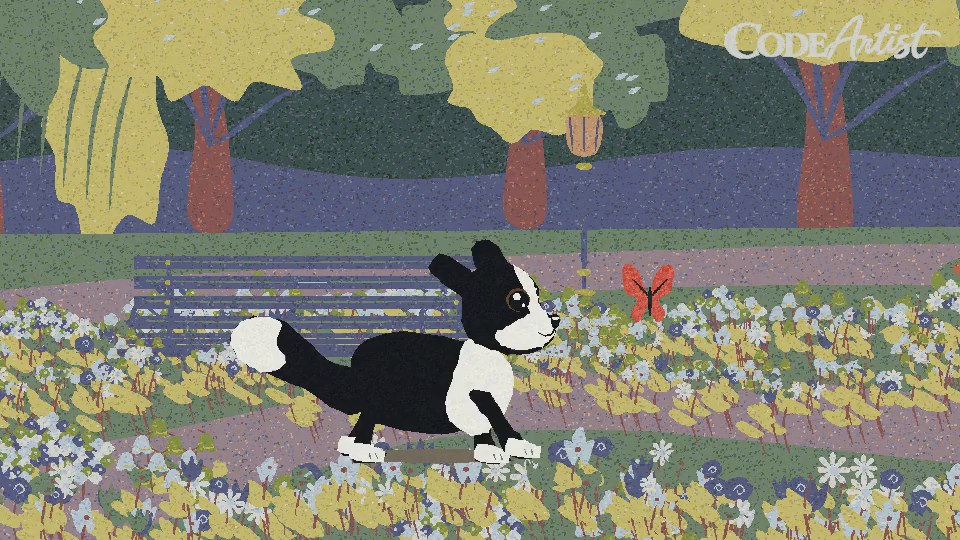}
\\[-1mm]
& \textsc{Riona} & & \textsc{Seurat}
\\[3mm]

\end{tabular}

\caption{
The same animation rendered in multiple \textit{CodeArtist} styles. 
Each pair shows a reference artwork and its corresponding first frame, with scene structure and motion preserved while the visual style adapts.
}
\label{fig:dog}
\end{figure*}

\textit{CodeArtist} supports plug-and-play style adaptation. 
Once a scene is implemented, it can be rendered in different artistic styles simply by switching the style program, without rewriting its underlying structure. 
In Figure~\ref{fig:flower}, the same flower construction is rendered with multiple styles, producing distinct forms, strokes, and textures rather than simple color variations.
Figure~\ref{fig:dog} applies the same workflow to animation, where a border collie running through the seasons is re-rendered in multiple styles while its subjects, layout, and motion remain unchanged.


\section{Conclusion}
We introduced \textit{CodeArtist}, a painting engine that represents artistic styles as reusable and executable programs while separating style from scene content and motion. 
Across six coding agents, \textit{CodeArtist} improved the average visual score from 34.93 to 38.42 while reducing generation time by 49.2\%  and cost by 49.1\%. 
It also maintained consistent styles across scenes, supported targeted visual revisions, and allowed lower-cost models to reuse drawing capabilities developed by stronger models. 
These results demonstrate that executable style programs provide a practical, controllable, and continuously improvable representation of artistic style.

\clearpage
\bibliography{main}
\bibliographystyle{plainnat}

\end{document}